\documentclass[lettersize,journal]{IEEEtran}

\usepackage[acronym]{glossaries}
\newacronym{AI}{AI}{Artificial Intelligence}
\newacronym{AOI}{AOI}{Area Of Interest}
\newacronym{BOA}{BOA}{Bottom-Of-Atmosphere}
\newacronym{CLAHE}{CLAHE}{Contrast Limited Adaptive Histogram Equalization}
\newacronym{COTS}{COTS}{Commercial Off-The-Shelf}
\newacronym{CNN}{CNN}{Convolutional Neural Networks}
\newacronym{CSC}{CSC}{Coarse Spatial Coregistration}
\newacronym{CPU}{CPU}{Central Processing Unit}
\newacronym{DL}{DL}{Deep Learning}
\newacronym{DNN}{DNN}{Deep Neural Networks}
\newacronym{EO}{EO}{Earth Observation}
\newacronym{ESA}{ESA}{European Space Agency}
\newacronym{FPGA}{FPGA}{Field Programmable Gate Array}
\newacronym{GIS}{GIS}{Geographic Information Systems}
\newacronym{GPU}{GPU}{Graphics Processing Unit} 
\newacronym{L0}{L0}{Level-0}
\newacronym{L1A}{L1A}{Level-1A}
\newacronym{L1B}{L1B}{Level-1B}
\newacronym{L1C}{L1C}{Level-1C}
\newacronym{L2A}{L2A}{Level-2A}
\newacronym{LEO}{LEO}{Low Earth Orbit}
\newacronym{LUT}{LUT}{Look-Up Table}
\newacronym{ML}{ML}{Machine Learning}
\newacronym{MSI}{MSI}{Multi-Spectral Instrument}
\newacronym{NIR}{NIR}{Near-InfraRed}
\newacronym{PyRawS}{PyRawS}{Python for RAW Sentinel-2 data}
\newacronym{S2-A}{S2-A}{Sentinel-2A}
\newacronym{S2-B}{S2-B}{Sentinel-2B}
\newacronym{SWIR}{SWIR}{shortwave infrared}
\newacronym{SSO}{SSO}{sun-synchronous orbit}
\newacronym{TOA}{TOA}{top-of-atmosphere}
\newacronym{VEI}{VEI}{volcanic explosive index}
\newacronym{VIS}{VIS}{visible}
\newacronym{VNIR}{VNIR}{Visible and Near-Infrared}
\newacronym{VPU}{VPU}{Visual Processing Unit}
\newacronym{CG}{CG}{Coarse Georeferencing}
\newacronym{B2B}{B2B}{Band-to-Band}

\usepackage{graphicx}
\usepackage{xcolor}
\usepackage{soul}
\usepackage[utf8]{inputenc}
\usepackage{url}
\usepackage{longtable}
\usepackage{amsmath}
\usepackage{hyperref}
\usepackage{amssymb}
\usepackage{booktabs}
\usepackage[normalem]{ulem}
\useunder{\uline}{\ul}{}

\usepackage[final]{changes}
\definechangesauthor[name={rev}, color=red]{rev1}
\definechangesauthor[name={rev}, color=green]{rev2}

\begin{document}
\title{Unlocking the Use of Raw Multispectral Earth Observation Imagery for Onboard Artificial Intelligence}

\author{Gabriele Meoni, 
Roberto Del Prete,
Federico Serva,
Alix De Beusscher,\\
Olivier Colin,
Nicolas Longépé

\thanks{Gabriele Meoni (corresponding  author, mail: gabriele.meoni@esa.int), Roberto Del Prete, Federico Serva, Alix De Beusscher, and Nicolas Longépé are with the $\Phi$-lab,  European Space Research Institute (ESRIN), European Space Agency (ESA), Via Galileo Galilei, 1, 00044 Frascati RM (Italy); Olivier Colin is with Copernicus \& Ground Segment / Data Management Division,  European Space Research Institute (ESRIN), European Space Agency (ESA), Via Galileo Galilei, 1, 00044 Frascati RM (Italy).}

\thanks{Manuscript received XX XX, XX; revised XX XX,XX .}}

\markboth{IEEE Journal of Selected Topics in Applied Earth Observations and Remote Sensing}%
{Meoni \MakeLowercase{\textit{et al.}}: Unlocking the Use of Raw Multispectral Earth Observation Imagery for Onboard Artificial Intelligence}

\IEEEpubid{XXXX--XXXX/XX\$XX.XX~\copyright~2021 IEEE}

\maketitle

\begin{abstract}

Nowadays, there is growing interest in applying Artificial Intelligence (AI) on board Earth Observation (EO) satellites for time-critical applications, such as natural disaster response. 
However, the unavailability of raw satellite data currently hinders research of lightweight pre-processing techniques and limits the exploration of end-to-end pipelines\replaced{, which extract}{ that could offer more efficient and accurate extraction of} insights directly from the source data.
To fill this gap, this work presents a novel methodology to \replaced{automate}{automatize} the creation of \deleted[id=rev1]{specific }datasets\added[id=rev1]{ for the detection of target events (e.g., warm thermal hotspots) or objects (e.g., vessels)} from Sentinel-2 raw data and other multispectral EO pushbroom raw imagery.   The presented approach first processes the raw data by applying a pipeline consisting of a spatial band registration and georeferencing of the raw data pixels. Then, \replaced[id=rev1]{it detects the target events by leveraging event-specific state-of-the-art algorithms}{it leverages the availability of state-of-the-art algorithms to detect the events of interest} on the Level-1C products, which are mosaicked and cropped on the georeferenced correspondent raw granule area. The detected events are, finally, re-projected back on the corresponding raw images.
We apply the proposed methodology to realize THRawS \added[id=rev1]{(Thermal Hotspots in Raw Sentinel-2 data)}, the first dataset of Sentinel-2 raw data containing warm thermal hotspots. THRawS includes \added[id=rev1]{1090 } samples containing wildfires, volcanic eruptions, and \added[id=rev1]{33335 }event-free acquisitions to enable thermal hotspot detection and general classification applications. This dataset and associated toolkits provide the community with both an immediately useful resource \replaced{to speed up future research}{as well as a framework and methodology acting as a template for future additions. With this work, we hope to pave the way to the research} on energy-efficient pre-processing algorithms and AI-based end-to-end processing systems on board EO satellites.

\end{abstract}

\begin{IEEEkeywords}
Onboard AI, raw dataset, Sentinel-2, Volcanic Eruption, Wildfire.
\end{IEEEkeywords}
%
\IEEEpeerreviewmaketitle

\section{Introduction}
\label{sec: introduction}
\IEEEPARstart{T}{he} ability of \gls{AI} to autonomously glean insights from remote sensing data has sparked research into its deployment onboard spacecraft \cite{furano2020towards}.
Furthermore, its application has proven effective in curtailing downlink data rates by preventive filtering of cloud-covered, corrupted data \cite{GiuffridaPhiSat, furano2020towards}, or with higher image compression rates \cite{guerrisi2022convolutional, guerrisi2023artificial}. 

Most of the previous works relying on \gls{AI} onboard \gls{EO} satellites were demonstrated on high-level satellite optical products \cite{di2022early,del2021board, ruzicka_ravaen_2022, mateo-garcia_towards_2021}, which exploit calibrated, ortho-rectified, and further processed images to lower instrumentation noise and other sources of radiometric distortion. Since such processing chains are designed for ground-based use, they are unsuited for onboard applications. 
The computational burden of the ground-based processing chain renders its implementation on small-satellites, --especially nano-satellites-- impractical. Furthermore, it is essential to note that this processing chain alters the raw signal and is susceptible to potential information degradation and loss. Finally, optimizing pre-processing steps is necessary to minimize the computational complexity, potentially reduce the requirement for extra onboard hardware, and enhance mission duty cycles.

In this scenario, a relatively limited number of works have explored the capability of \gls{AI} models to directly process raw data with minimal pre-processing to reduce the need for computationally intensive onboard operations \cite{del2023first, fanizza2022transfer}. This is primarily due to the scarcity of literature containing datasets made of raw data and methodologies to produce such datasets for applications relevant to onboard \gls{AI} processing. \added[id=rev1]{Indeed, increasing the availability of datasets comprising raw imagery could potentially diminish the reality gap prevalent in contemporary studies. This could be achieved by facilitating the training of \gls{AI} models on satellite data, processed via methodologies that are congruent with the constraints of onboard power and memory capacities. Furthermore, this could stimulate research into more efficient processing techniques. These lightweight techniques could offer enhanced trade-offs between energy consumption, processing time, and output quality, as compared to solutions that were initially conceived for terrestrial applications.} \IEEEpubidadjcol

\replaced[id=rev1]{This work addresses the limited availability of raw imagery for multispectral pushbroom imagery. To this aim, we propose }{ To fill this gap, this paper proposes} for the very first time a methodology to datasets for the detection of ``events" or target objects (e.g., thermal anomalies, vessels) applied to Sentinel-2 raw data and that can be \replaced[id=rev1]{excited}{excted} to other multispectral \gls{EO} pushbroom raw imagery. \added[id=rev1]{To this goal, the proposes approach capitalises on existing datasets and algorithmic solutions designed for \gls{L1C} products to extract the identify the corresponding raw images containing the target events and speed-up the dataset creation.}

In the case of Sentinel-2 data, ``raw'' refers to decompressed Sentinel-2 \gls{L0} data with additional metadata, which better represents the ``raw" data produced by the \replaced[id=rev1]{multi-spectral}{multispectral} sensor.

To automate and make our methodology fully reproducible, we implemented an open-source Python package named \textit{\gls{PyRawS}}\footnote{PyRaws, GitHub repository. Available online at: \url{https://github.com/ESA-PhiLab/PyRawS}. Last accessed on 30/03/2024.} that includes utilities for Sentinel-2 raw and \gls{L1C} data processing.   

\begin{figure*}[t]
    \centering
    \includegraphics[width=1\textwidth]{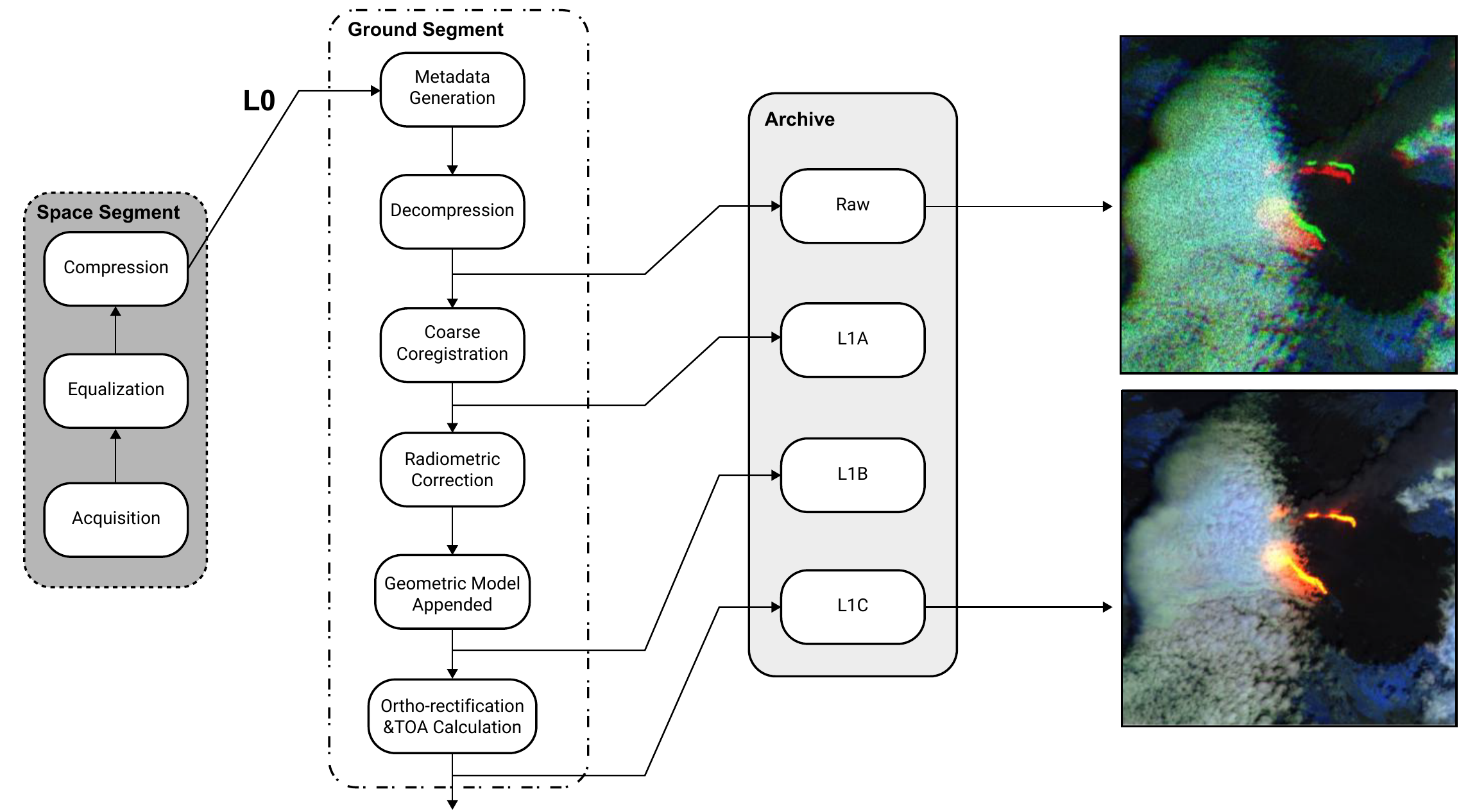}
    \caption{Illustration of the processing chain from satellite data to L1C data. 
    The images in the raw and L1C format show an eruption of the Etna volcano, Italy, for the two processing levels as RGB-like images.}
    \label{fig: L0-L1C}
\end{figure*}

To showcase the validity of the proposed methodology, we applied our methods for the realization of THRawS (Thermal Hotspots in raw Sentinel-2 data), a dataset comprising a collection of Sentinel-2 raw data for detecting warm temperature hotspots. 
We considered this specific application because of its potential to enhance early warning systems and prevent fires from spreading \cite{del2021board, 9984159}, contributing to the safety and security of people in high-risk areas and critical infrastructures, e.g., power lines and oil pipelines. 

Since the objective of using raw data is to foster the research of lightweight pre-processing methods, \replaced[id=rev1]{for the implementation of THRawS we leveraged a lightweight \gls{CSC} approach}{we leveraged a lightweight \gls{CSC} approach for the implementation of THRawS}. Such \gls{CSC} method has the potential to be applied onboard because of its timing and accuracy trade-offs.  

The remainder of the paper is organized as follows. 
Section \ref{sec: sota} provides a State-of-the-art analysis of the processing pipelines designed for multispectral pushbroom imagery, differentiating between 
missions performing processing on the ground and on board. Moreover, an \replaced[id=rev1]{insight}{insigth} on the 
availability of raw multispectral \deleted[id=rev1]{raw }data and their use in current missions leveraging \gls{AI} onboard \gls{EO} data is provided. 
Section \ref{sec: DatasetCreation} overviews the proposed dataset creation methodology.
Section \ref{sec: THRawSCasetStudy} showcases how we applied the proposed methodology to design THRawS. 
Section \ref{sec: Results} presents the results in terms of the number of thermal hotspot patches included in the THRawS dataset. Moreover, it provides a description of their spatial and time coverage.\added[id=rev1]{ }
Section \ref{sec: Discussion} discusses our results and the suitability of using raw data for future usability for onboard \gls{AI} applications. Finally, Section \ref{sec: Conclusions} draws the conclusion of this work.
Finally, we provide more details on the Sentinel-2 mission and a comparison in terms of accuracy and latency of the \gls{CSC} technique adopted to design THRawS to other \gls{B2B} alignment solutions in the Appendix.

\section{Background and Motivation}
\label{sec: sota}

\begin{figure*}[tb]
    \centering
 \includegraphics[width=0.97\linewidth]{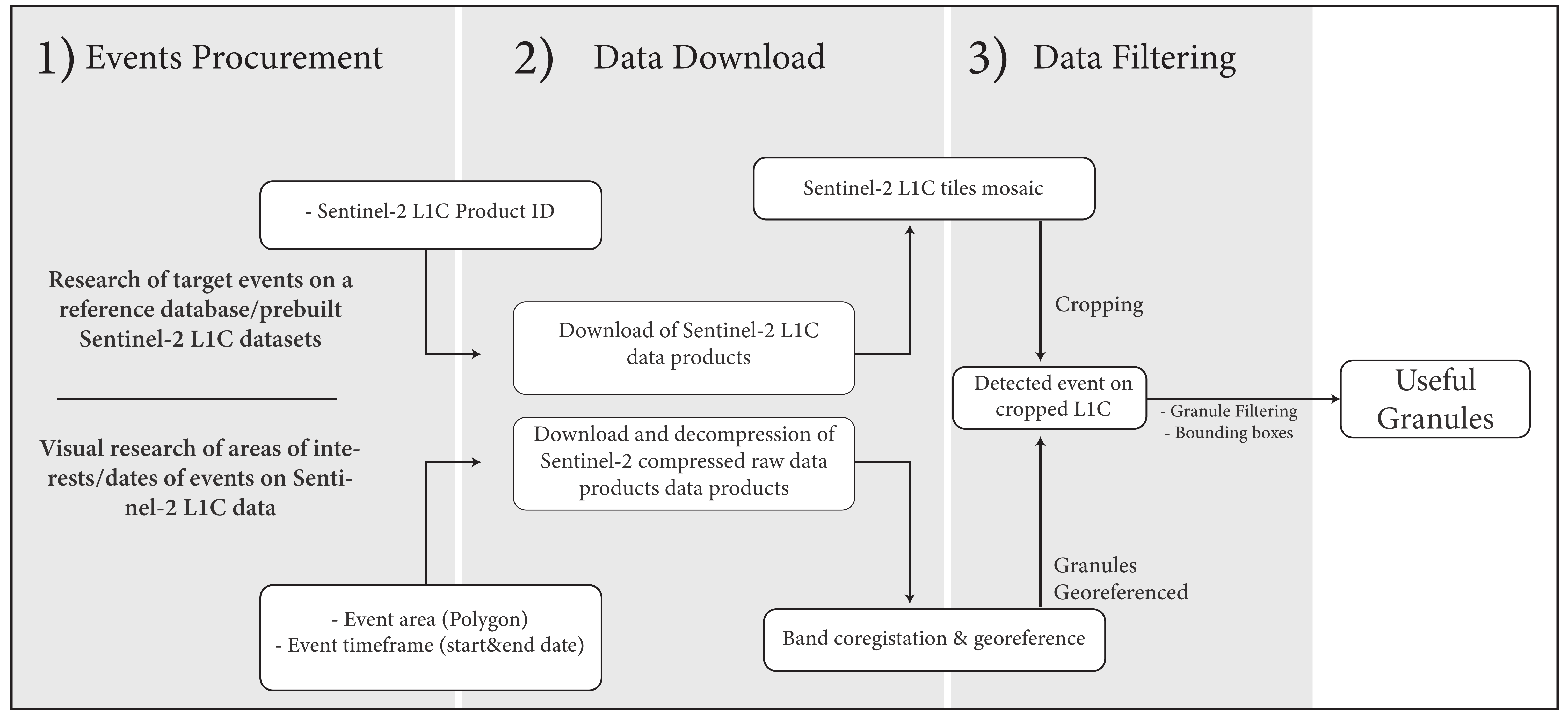}
    \caption{Overview of the dataset creation methodology consisting of three main steps: 1) procurement of the list of events by visual inspections of other existing databases, 2) data download, 3) filtering of useful granules by using state-of-the-art algorithms designed for \gls{L1C} data, which are mosaicked and cropped on the areas of the raw granules of interest.}
    \label{fig: datasetcreationMethodology}
\end{figure*}

\subsection{Multispectral data processing chain}
\label{subsec: S2dataProcessing}

In missions such as Sentinel-2 \cite{drusch2012sentinel} or Landsat-8 \cite{wulder2019current}, significant post-processing is applied to enhance satellite multispectral pushbroom imagery prior to their use for the creation of high-end applications\replaced[id=rev1]{-}{ }specific products. Typical correction steps include radiometric, geometric and orthorectification algorithms \cite{richards2022remote}; most of these steps are applied at ground-segment level.
Since our dataset methodology targets Sentinel-2 imagery, this section dedicates particular focus to Sentinel-2 imagery. However, \replaced[id=rev1]{although}{despite} the implementation of such processing solutions and product\deleted[id=rev1]{s} nomenclature differ among various missions, fundamental operations such as band\deleted[id=rev1]{s} coregistration, image georeferencing, radiance to \gls{TOA} reflectance conversion \replaced[id=rev1]{are common steps for many of these data processing chains}{are common processing steps for many of these missions data processing chain} \cite{S2PSD}\footnote{Landsat 8 (L8)
Data Users Handbook. Available online at: \url{https://d9-wret.s3.us-west-2.amazonaws.com/assets/palladium/production/s3fs-public/atoms/files/LSDS-1574_L8_Data_Users_Handbook-v6.0.pdf}. Last accessed on 28-03-2024.}.
In the Sentinel-2 processing chain, images are downloaded and processed to be distributed to the end-users through a processing chain whose schematic representation is depicted in Fig. \ref{fig: L0-L1C}.
Data sensed during a given acquisition are organised into granules, each one encompassing  
the area captured by each detector in a 3.6s interval \cite{S2PSD}. 
Since the satellite operates in pushbroom mode, a granule encapsulates data from all the bands a sensor acquires. Data produced by the sensor are equalized and compressed onboard before the downlink.  

After the download, data available on the ground are defined as \gls{L0}. The latter are, first, processed to produce metadata, including geographical information, a quick look, and other ancillary information. Then, with their additional metadata, \gls{L0} data are decompressed. 

In the frame of this paper, the data obtained at this point of the processing, including metadata, are defined as ``raw data". 
\gls{L1A} products are then generated by applying a coarse spatial band coregistration process to the raw data.
This is followed by processing \gls{L1A} data to compute radiometrically corrected radiances through various processing steps, including --among others-- the inversion of onboard equalization. In addition,  the radiometric corrected geometric model is refined and appended to the radiometric corrected \gls{L1A} data to produce \gls{L1B} products.

Lastly, \gls{L1C} data are derived from \gls{L1B} products through a process of geometric correction,  
which includes sub-pixel multispectral spatial registration, ortho-rectification, and calculation of \gls{TOA} reflectances. Cloud and land masks are also generated within this step. Various geometric and radiometric periodic calibration activities are performed to ensure high-quality products, as detailed in \cite{S2Calibration}. 

Note that \gls{L1C} products are delivered for predefined $100\times100 \: km^2$
\emph{tiles}, and a given acquisition may have partial or complete data coverage of a certain \gls{AOI} in a certain raw data granule. 
Therefore, in general, multiple \gls{L1C} tiles shall be mosaicked to cover the whole raw \gls{AOI}.

Since the inversion of the onboard equalization is performed only at \gls{L1A} data level, 
its effects are still present for raw data. Similarly, as previously mentioned, the equalized satellite data
has undergone on board a process of equalization before compression \cite{drusch2012sentinel}, which aims at minimizing the effect of the wavelet-based compression algorithm on the accuracy/integrity of the detector measurements provided after decompression on the ground. The effects of onboard compression cannot be compensated because of the lossy nature of the \replaced[id=rev1]{used compression scheme}{compression scheme used}. Consequently, eventual distortions due to onboard compression and calibration represent the main difference between raw and data produced by the sensor.
 Note that the products in the chain from \gls{L0} to \gls{L1B} included have not been available to users at the time of writing,
while \gls{L1C}, atmospheric-corrected \gls{BOA} Level-2A data as well as higher level derived products for different domain applications, such as marine ecosystem 
monitoring\footnote{\url{https://shorturl.at/jvGR6}; 
last accessed on 2023-12-19.} and land cover/land use 
assessment\footnote{\url{https://seom.esa.int/page_project025.php}; last accessed on 2023-12-19.} 
are available from Copernicus and ESA projects. However, the release of \replaced{lower-end Sentinel-2 products to the public is foreseen.}{additional Sentinel-2 raw imagery to the ones provided in this manuscript and other previous work to the public is planned}. 
Products including and beyond \gls{L2A} are out of the scope of this study, which focuses on the combined use of raw and \gls{L1C} data to create a dataset made of raw data.

\subsection{Multispectral processing chains for EO missions leveraging onboard ML}

The application of \gls{AI} on board satellited constitutes an established scientific and commercial venture. In particular, the remote sensing community has heightened the potential to directly handle unprocessed data through the techniques of \gls{ML} \cite{danielsen2021self} and \gls{DL} \cite{del2021board, GiuffridaPhiSat}. 
Prior to the application  of \gls{ML} models, current and upcoming missions predominantly favor the application of pre-processing schemes as a prerequisite  \cite{helber2019eurosat,del2021board, ruzicka_ravaen_2022, mateo-garcia_towards_2021}. Such solutions usually consist of streamlined workflows that apply only a simplified subset of geometric and radiometric corrections  of the ones performed on the ground. 

An illustrative example is the case of the 6-U $\Phi$sat-1 \cite{GiuffridaPhiSat} CubeSat, the first satellite inferring a \gls{CNN} on a \gls{COTS} edge device, i.e., the Intel-Movidius Myriad 2 \gls{VPU}. 
Its target application was to perform onboard pixel-level cloud detection by processing three selected bands of the hyperspectral cube. 
The image processing workflow involves creating the hyperspectral data cube and undertaking \gls{B2B} spatial registration on three selected bands.
Another example is the HYPSO-1 (HYPer-spectral Smallsat for ocean Observation), a CubeSat equipped with \added[id=rev1]{a} hyperspectral payload targeting the observation of ocean color to detect algal blooms along Norwegian costs \cite{danielsen2021self}. The hyperspectral image processing chain on board HYPSO-1 uses the methodology described in \cite{8921350}, which implements radiometric and geometric corrections as linear operations. 

A more complex processing chain is planned for the upcoming $\Phi$sat-2 mission \cite{melega2023implementation}. $\Phi$sat-2 imager is providing seven multispectral bands covering \gls{NIR} and visible ranges with 4 m spatial resolution and a panchromatic band (sub-meter spatial resolution). The payload processing chain processes raw data and provides three processing level: \gls{L1A}, consisting of unregistered and no-georeferenced radiance data, \gls{L1B}, which are fine-georeferenced and fine registered radiance data (< 10 m root mean square error), \gls{L1C}, which are registered and geo-referenced \gls{TOA} reflectance data. Differently from Sentinel-2 data, \gls{L1C} data are not ortho-rectified. Given the unaivability of $\Phi$sat-2 imagery, emulated sensory data at \gls{L1A}, \gls{L1B}, \added[id=rev1]{and} \gls{L1C} levels were provided to the competitors of the \textit{Orbital AI} challenge\footnote{Orbital $\Phi$sat-2. Available online at \url{https://platform.ai4eo.eu/orbitalai-phisat-2}. Last accessed 30/03/2024.} \cite{LongepeOrbitalAI} to investigate different applications for onboard \gls{AI}.
One of the main novelty of $\Phi$sat-2 and Orbital AI is the investigation of lower-end products for onboard \gls{ML} applications. However, no emulation of raw data is provided to the end users nor a use of raw data is foreseen in the mission concept of operation. 

As stated above, current approaches engage data processing with the intention of reconstructing higher-level products onboard, thereby not directly exploiting raw products. 
The deficiency of raw datasets and the unavailability of methods to treat them inadvertently ease the pathway for higher-level products onboard processing.
Nonetheless, the adoption of such an approach contributes considerably to the computational cost within the processing workflow and introduces temporal latencies. Indeed, following the empirical assessment of the inference time-lapse on various hardware accelerators such as NVIDIA Jetson Nano, Google Coral, and Myriad X, Mateo-Garcia et al. \cite{mateo2022orbit} asserted that the time overhead associated with the initial processing stages of Sentinel-2, encompassing \gls{B2B} alignment\added[id=rev1]{--i.e., spatial registration of a band with respect to a reference band --}, radiometric calibration, and granular coregistration, merits a scrupulous evaluation.

To the knowledge of the authors, the first work to directly address the problem of using raw data in an end-to-end fashion for onboard \gls{ML} applications was conducted in the frame of ``the OPS-SAT case"  competition \cite{derksen2021few, meoni2024ops}. This data-centric competition hosted on the \gls{ESA} Kelvins platform was investigating few-shot learning for onboard satellite applications \cite{KelvinsOPSSAT}. 
The competition was based on the images produced by the OPS-SAT satellite, equipped with an $80m \times 80m$ resolution RGB via on-sensor Bayer Pattern camera \cite{OPSSATSpecifications}. In this case, only one frame is acquired from a single snapshot by placing a color filter on top of every sensor pixel. A so-called demosaicing or debayering process is applied to reconstruct a full-color image from a sampled one-color pixel camera. 
Thereby, images were not affected by misregistration errors due to multispectral \gls{B2B} alignment processes, obviating one of the most critical pre-processing constraint for the data stack's formation. 

Similarly, the work by Fanizza et al. \cite{fanizza2022transfer} investigates the effect of band\deleted[id=rev1]{s} misalignment for ship detection on 768 × 768 pixels RGB images from the Airbus Ship Detection dataset was used for both training and testing. Because of the lack of raw data, the band misaligment was emulated. 

At the best of the our knowledge, our previous work \cite{del2023first} is the first work investigating the feasibility of using multispectral raw imagery. To this aim, our previous work \cite{del2023first} relies on a dataset containing Sentinel-2 raw coarsely-registered granules to perform onboard vessel detection through an end-to-end pipeline. Despite the preliminary findings of the work show that processing raw data in an end-to-end fashion is possible, the difficulty of extending these results for different applications is raised by the lack of a methodology to automate raw dataset creation and labelling. Indeed, the dataset used for our study was manually labelled by using a Sentinel-2 \gls{L1C} dataset as reference. Because of that, the methodology propose hereby extends our previuos work by enabling an automated labelling of raw data by using annotations on the correspondent \gls{L1C} tiles.

Despite not-addressing this problem specifically, the need for raw data availability also emerges from the work \cite{mateo2022orbit}, which describes the WildRide mission. The latter aims to illustrate that the onboard application of \gls{AI} is instrumental in minimizing latency, thereby accelerating the delivery of essential flood mapping information. The \gls{DL}-based model utilized to perform flood detection was trained on RGB data provided by a miniaturized camera. 

Nonetheless, given the lack of raw data, Worldfloods was initially trained on Sentinel-2 \gls{L1C} images by using all 13 bands. As a result, Worldfloods faced challenges arising from the data-shift problem due to the significant difference in terms of radiometric and spectral resolution between the Sentinel-2 sensor and the onboard RGB imager. Therefore, the authors retrained the model on four acquisitions from the used RGB camera to obtain acceptable performance on the raw uncalibrated RGB images, which were manually labelled after the satellite's deployment. 
This underlines the importance of having raw satellite datasets for the training of \gls{AI} models to take into account: a) the impact of the lack of calibration, b) domain-gap problems, c) misregistration errors, and d) other non-ideal effects affecting raw data.

%


\section{Dataset Creation Methodology}
\label{sec: DatasetCreation}
\begin{figure*}[t]
    \centering
\includegraphics[width=0.95\linewidth]{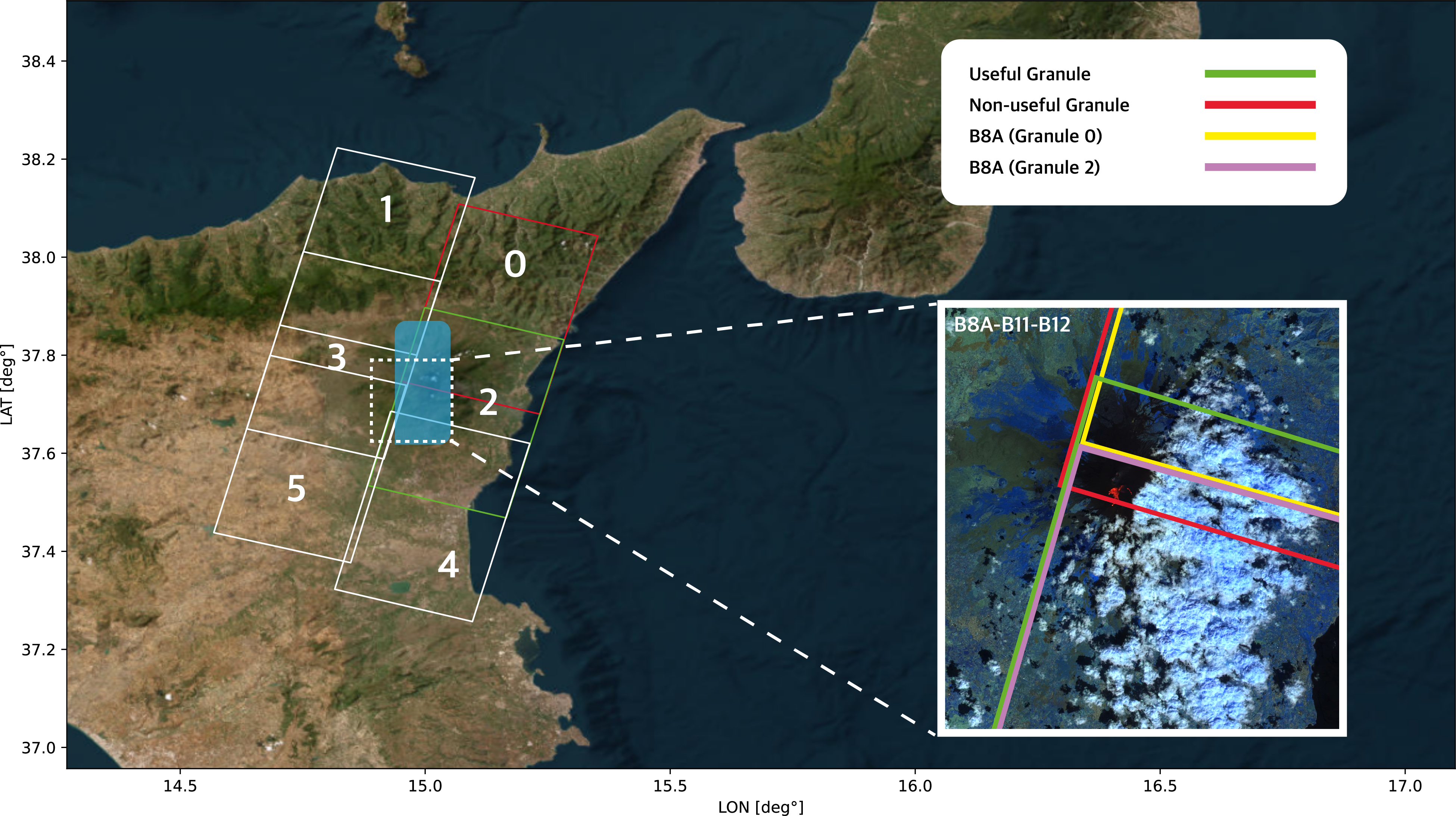}
    \caption{Example showcasing the downloaded granules for a specific event of the THRawS dataset.
    The polygon used for data retrieval (light blue) intercepts several granules marked in red, green, and white.
    The volcanic eruption is included only on the granules whose boundaries are marked in green and red. On the right, a zoomed view shows the band $B_{8A}$ (the first of the collection) of the green and red granules in yellow and pink. Since the band $B_{8A}$ of the green granule only includes the volcanic eruptions, the green granule is the only ``useful granule''. On the contrary,  despite the red polygon partially surrounds the event of interest, this happens for other bands than $B_{8A}$. Because of that, the event is ``non-useful'' for our definition of the band collection $B_S 0 = [B_{8A},B_{11}, B_{12}]$. }
    \label{fig: needForFiltering}
\end{figure*}
%
%
\label{subsec: MethodologyOverview}
The process of creating a Sentinel-2 raw dataset for the specific purpose of detecting ``events" or target objects can be comprehensively understood through the graphical representation depicted in Fig. \ref{fig: datasetcreationMethodology}\replaced[id=rev1]{, which}{ that} showcases all \deleted[id=rev1]{the }three steps required to create the dataset. 
The first step is procuring a list of events from online databases or other sources. Each event in the list shall correspond to one or more raw granules. 
To download such raw data, it is necessary to specify a polygon marking the \gls{AOI} surrounding the event and a range of dates (start acquisition - end acquisition). Such a polygon used to specify the raw granules is created by visually inspecting the \gls{L1C} products. 
\replaced[id=rev1]{This}{These} procedure leads to the download of all the compressed raw granules whose reference band ($B_{02}$) intersects the polygon and whose sensing date is included in that range.

Appropriately selecting the area of the polygon requires considering that for a specific point, the entire collection of the bands could be located in different granules before the spatial registration.
Therefore, to collect all the bands for the events of interest, one can use a polygon area of $28 \times k$ $km^2$ centered at the coordinates of the events, where $k$ is application-specific, which is equal to the double of the maximum bands parallax\replaced[id=rev1]{e}{i}s for a single acquisition.
For what concerns the parameter $k$, it shall be higher than the maximum West-East size of the target events.
A size of $14km$ corresponds to the maximum along-track distance commonly covered by all spectral bands of \gls{MSI}. 
Therefore, since the satellite track approximately extends from north to south with a \replaced[id=rev1]{smaller}{more minor error} closer to the Equator, selecting a polygon that extends $14km$ vertically toward north and south will ensure that all the bands for the center of the event will be downloaded. 

Once the first step is complete, it is possible to download the raw granules and all the related \gls{L1C} products intersecting the download polygons for each granule.  
After downloading, the compressed raw data shall be decompressed and reformatted as TIFF files. The download and the decompression of the interest data lead to the conclusion of the second step. 

However, the granules downloaded through \replaced[id=rev1]{this}{these} procedure generally cover an area higher than \gls{AOI}.
Indeed, since the polygon is not exactly oriented along the satellite track line and because the real \gls{AOI} could be much smaller than the polygon, this procedure generally leads to a higher number of raw data granules than required. 
This can be seen in the left image of Fig. \ref{fig: needForFiltering}, which shows the granules downloaded (red rectangular shapes) for the volcanic eruption of Etna on 30/08/2021 included in the THRawS dataset. The polygon used to specify the \gls{AOI} is shown in \replaced[id=rev1]{light blue}{white}. But, clearly, only the raw data granules in yellow and pink actually contain the events. Therefore, there is a need to identify and label the granules containing the events of interest. 

To this aim, we conceived a methodology that filters the raw data granules containing the events of interest by processing \gls{L1C} tiles by relying on the availability of state-of-the-art algorithms for \gls{L1C} products. 

Such state-of-the-art algorithms usually exploit a subset of the 13 bands included in Sentinel-2 data. 
Let us define as $B_S \triangleq [B_x, B_y, ..., B_z]$ the ordered list of the bands used by the identified algorithm, where $B_x$ is the first band of the collection.
Then, for each event in the dataset, every granule is processed as follows. First, we perform the coregistration of the different bands of $B_S$ with respect to $B_x$. 
Spatial coregistration algorithms are designed to eliminate displacement between bands, errors in skewness, rotation, and warping \cite{kennedy2003automated}.
In our case, this step is mainly necessary since the bands in a granule feature a displacement due to the pushbroom nature of the sensor and additional non-ideal effects \cite{S2PSD}. Such band shifts will make it difficult to retrieve and visually inspect events on raw data after being identified. 
By applying the band spatial coregistration with respect to $B_x$, among other effects, all the bands of $B_S$ will be shifted along and across track to match the area covered by the band $B_x$. The missing pixels due to the shift procedure could be cropped, filled by zeros, or by pixels of adjacent granules when available. 
After applying the band coregistration, we performed geo-referencing of the first band in the granule. This step is needed since only the coordinates of the corners of the entire granule footprint are provided in the granule metadata, but not for the different bands \cite{S2PSD}. 

Once the band $B_x$ raw granule has been georeferenced, it is possible to crop \gls{L1C} data on the same area of its band $B_x$. To this aim, all the \gls{L1C} tiles intersecting the polygon used to download raw data are mosaiced and cropped on the same area of the band $B_x$. To this aim, the information provided by $B_x$ coordinates as a common reference for \gls{L1C} and raw data.

After cropping, these adjusted \gls{L1C} tiles are processed using the identified reference algorithm to search for events of interest. One or more bounding boxes are created on the \gls{L1C} data for each event.
When finding a bounding box is impossible, the correspondent granule is discarded. Otherwise, the bounding boxes are warped back on the granule using an affine transformation and manual fine-tuning. 

We mark \textit{useful granule} as a granule whose band $B_x$ includes at least one bounding box. It is necessary to remark that this definition of \textit{useful granule} depends on the band collection used. In the case of the volcanic eruption ``Etna\_00" of the THRawS dataset shown in Fig. \ref{fig: needForFiltering},  the volcanic event is included in both the granules 0 and 2, respectively marked in red and green. However,  since for the THRawS dataset  $B_x = B_{8A}$, the procedure previously described will lead to selecting only granule 2 as a \textit{useful granule}. Indeed, when considering the area covered by the band $B_{8A}$ in granule 0 (yellow rectangle), one can see that it does not include the volcanic eruption.

\section{The THRawS Case Study}
\label{sec: THRawSCasetStudy}

 This section demonstrates how we applied the proposed methodology described in Sec. \ref{subsec: MethodologyOverview} to create the THRawS \deleted[id=rev1]{(Thermal Hotspots in Raw Sentinel-2 data)}.

As mentioned in Sec. \ref{subsec: MethodologyOverview}, the first step to creating a dataset according to the proposed methodology is to identify the event's location and time ranges by relying on existing datasets or online databases.
More specifically, THRawS contains volcanic eruptions and wildfire events, for which we identified two different sources. 

The starting point for selecting tiles with eruptions is the Smithsonian Institution
database \cite{VolcanoDatabaseWeb}, providing detailed information on global volcanism.
From the database, we selected only events occurring during the \gls{S2-A} launch until 
the time of writing, i.e., from 2016 to 2022, and we aimed to sample different latitudes and seasons.
From the database, we only selected more intense and explosive eruptions, with 
\gls{VEI} larger than 1, since effusive eruptions are 
less dangerous \cite{abbott2008natural}.
As events can last for several years, we initially selected up to three volcanic events for each 
eruption and up to three eruptions for a single location to ensure the diversity of the dataset.  However, during the revision of the dataset, some raw data granules originally marked as not-events were found to contain volcanic thermal hotspots, leading to more than three volcanic events for some of the eruptions.

For shorter (days-months) eruptions, we selected acquisitions as close as possible
to the event starting date to test the \replaced[id=rev1]{detection capability}{capability of detection} in the early stages. 
From the list of eruptions, we visually inspected the candidate \gls{L1C} tiles
to confirm that the event was captured.
For each of the selected \gls{L1C}, we marked the location of the volcanic event. We used its coordinates as a center of the rectangular polygon to download raw data granules, as detailed in Sec. \ref{subsec: MethodologyOverview}. In particular, we selected the horizontal size of the download polygon to be $k = 10 km$ since it was sufficient to contain the broadest warm temperature hotspot in the dataset. 

Compared to volcanic eruptions, fire events are challenging to detect as they are shorter-lived and their 
locations are not known in advance compared to volcanic eruptions.
We compiled a list of events starting from the Copernicus Emergency Monitoring 
System database\footnote{\url{https://emergency.copernicus.eu/mapping/list-of-activations-rapid}; last accessed on 2023-12-19.} 
and integrated the resulting list of events with real-time detection 
services\footnote{\url{https://firms.modaps.eosdis.nasa.gov/map}; last accessed on 2023-12-19.}
and information available online from space and environmental agencies.
As for eruptions, candidate \gls{L1C} acquisitions have been visually inspected to verify the presence of a firefront or extensive smoke. Finally, we retained a number of fire events roughly equal to that of eruptions to balance the dataset distribution.

\added[id=rev1]{Both wildfire and volcanic eruptions events were selected by visual inspection independently and, then, crosschecked by two experts to ensure reliability and mitigate the effect of human biases.}
In addition to granules containing \gls{L1C} data, we specifically introduced hard negatives to optimize model performances by a proper selection of Not-events. This approach aims to establish consistent relationships between thermal anomalies and network outputs as it enables the identification of specific anomalies in the image not necessarily related to volcanic eruptions, e.g., crater rim edges. Selecting not-events involves visually inspecting each granule and manually selecting to ensure the absence of hotspots. 
It is worth noting how \replaced[id=rev1]{this is important to allow }{important this is in allowing} for a more accurate ingestion of the data in \gls{DNN} models. 

After selecting the reference warm temperature hotspots events, we proceeded with the download and filtering of the corresponding raw and \gls{L1C} data. 

In particular,  we selected $B_S=[B_{8A}, B_{11},  B_{12}]$, given the choice of the reference algorithm to detect warm thermal hotspots in thermal anomalies by Massimetti et al. \cite{massimetti2020volcanic}. 

The next required step is the coregistration of the different multispectral bands in the bands collection $B_S$. 
Indeed, given two generic bands $B_n$ and $B_m$ of a granule, the band $B_n$ is generally shifted of a certain number of pixels $S_{{B_n}\_{B_m}}$ with respect to $B_m$. 
In general, it is possible assume that $S_{{B_n}\_{B_m}}$ is made of two components as shown in Eq. \ref{eq: snm}:
\begin{equation}
    S_{{B_n}\_{B_m}} = \overline{S_{{B_n}\_{B_m}}} + {\Delta}S_{{B_n}\_{B_m}}
    \label{eq: snm}
\end{equation}
where $\overline{S_{{B_n}\_{B_m}}}$ is the systematic shift due to pushbroom acquisition mode and additional offset \cite{S2PSD}, whilst ${\Delta}S_{{B_n}\_{B_m}}$ is an aleatory component due to mechanical vibrations and other non-ideal effects. 

Spatial coregistration algorithms are generally designed to eliminate both the components of $S_{{B_n}\_{B_m}}$ so that the spatial displacement between $B_n$ and $B_m$ can be minimized \cite{anuta1970spatial, nandy2004edge} in addition to errors in skewness, rotation, and warping \cite{kennedy2003automated}.
To achieve these objectives, as detailed in Section \ref{sec: sota}, spatial coregistration techniques rely on a suite of algorithms that are generally computationally intensive and could require dedicated additional hardware to be implemented on board satellites \cite{GiuffridaPhiSat, anuta1970spatial, nandy2004edge, Guoxia}. 
Since one of the objectives of this work is to foster the study of lightweight and energy-efficient pre-processing algorithms, to design the THRawS, we adopted a coregistration solution designed to compensate only the systematic shift error component $\overline{S_{{B_n}\_{B_m}}}$ by shifting $B_m$ of a fixed value equal to $\overline{S_{{B_n}\_{B_m}}}$. 

To this aim, as for \cite{8921350}, our solution uses one-time pre-calculated spatial shift values to compensate for the average spatial displacements both along and across the satellite track due to the sensor's pushbroom nature and additional sensor offsets. 
An example of this pre-computed calculation is reported in Table \ref{tab:CoregError} in the Appendix.
The thus developed approach does not require using ground control points nor other ancillary data. In addition, it does not require to perform keypoints extraction and matching, typical of features-based technique \cite{ordonez2021comparing, lowe2004distinctive,lowe2004distinctive, bay2008surf, leutenegger2011brisk, rublee2011orb}.
These shift values do not consider effects due to noise, attitude disturbances, or depending on the surface characteristics. However, as shown later in Table \ref{tab:CoregError}, for Sentinel-2 data $\overline{S_{{B_n}\_{B_m}}} >>  {\Delta}S_{{B_n}\_{B_m}}$, especially for the along-track component (\textit{along-track shift}). 
Owing to its limited precision, we term our coregistration technique ``coarse". 
As outlined in Appendix, the average registration errors remain smaller than the smallest identified thermal hotspots. Because of that, the use of the proposed \gls{CSC} does not lead to loss of events.  
In addition, the proposed approach is lightweight, which makes it promising for onboard satellite applications and accelerates the dataset creation. The methodology to estimate the shift values used for the \gls{CSC} is detailed in Appendix.

After applying the band coregistration, we adopted a custom band \gls{CG} scheme based on the fixed band-to-band dispacement values used for the \gls{CSC} applied to the granule corner coordinates to determine the coordinates of each band's corners. Being a lightweight solution, this approach is promising to be adopted for onboard satellite applications. However, it assumes that the coordinates of the four granule corners are known. This information is contained in the granule metadata and requires additional processing steps to be made available on board, which were not investigated in this study. 

\begin{figure}[b]
    \centering
    \includegraphics[width=0.8\linewidth]{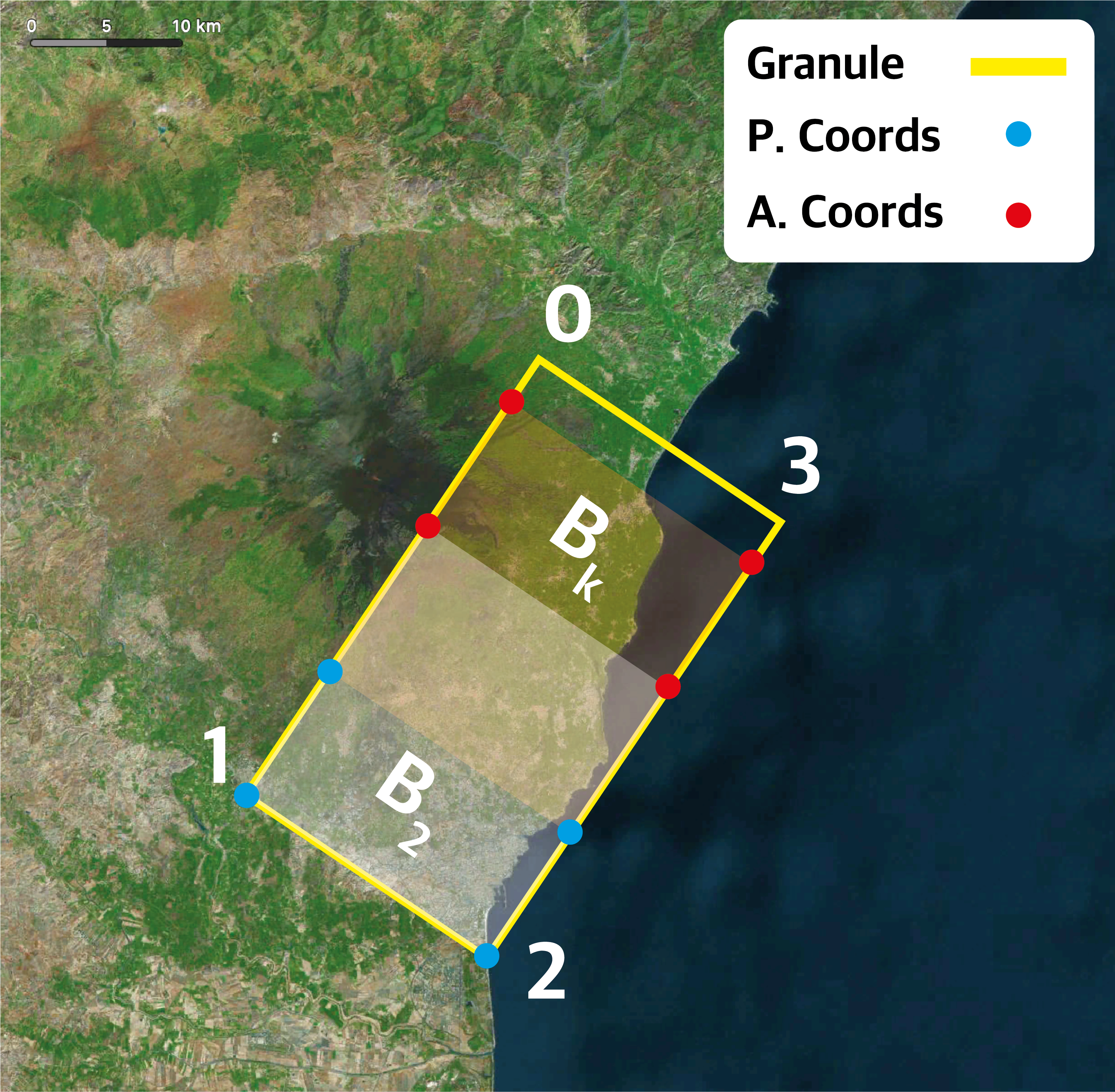}
    \caption{Pictorial view of a granule with Prior and Afterwards coordinates}
    \label{fig:geo-granule}
\end{figure}

Although we applied the following methodology to band $B_{8A}$ only, our approach is general, and, therefore, we provide a description for a generic band $B_k \in B_S$. 
We initiate this process by extracting the coordinates of the four corners of a granule, which constitute the essential ancillary information provided in the metadata of the Sentinel-2 raw product.
Referring to Fig. \ref{fig:geo-granule}, we define two key variables for our methodology. The first, Prior Coordinates ($PC$), represents the initial two corner coordinates scanned by the pushbroom system. The second, Afterward Coordinates ($AC$), denotes the final two corner coordinates scanned. 

\begin{figure*}[!tb]
    \centering
    \includegraphics[width=\linewidth]{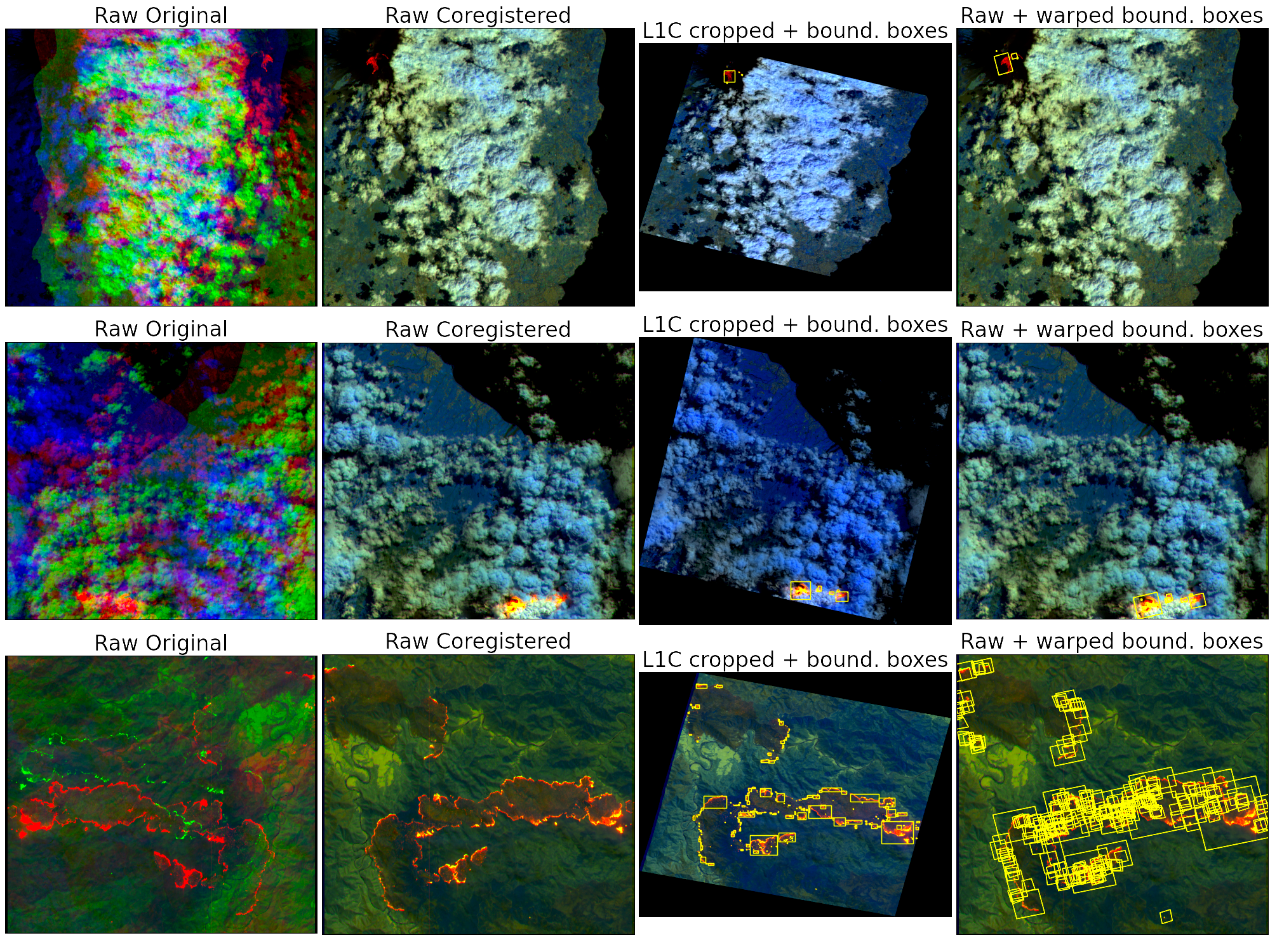}
    \caption{Each row shows all the processing steps for the bands $B_{8A}, B_{11}, B_{12}$ of the various raw data granules: the first image from left of each row displays the raw data granule, the second image shows a spatially registered granule, the third image shows the correspondent \gls{L1C} tiles cropped on the coordinates of first band of the raw data granule band collection with the detected bounding boxes, and the rightmost image showcases the bounding boxes warped on the raw data granules.}
    \label{fig: DatasetsExamples}
\end{figure*}

Given the known location of Prior Coordinates for band B02 ($PC_{B02}$), \replaced[id=rev1]{our method}{the method we have devised} utilizes the coarsely estimated offsets $S_{{{B_k}}\_{B_{02}}}$ from this reference band to determine the Prior Coordinates $PC_{B_k}$ for any arbitrary band $B_k$.  

For a band $B_k$, we can estimate the offsets from the reference band using the following equation:

\begin{equation}
    PC_{B_k} = PC_{B02} + |S_{{{B_k}}\_{B_{02}}}| \cdot \frac{R_{B_k}}{R_{B_{02}}} \cdot \frac{AC_{B_{09}}}{G_L}
\end{equation}

Here, $S_{{{B_k}}\_{B_{02}}}$ represents the shift value between band $B_k$ and the reference band $B02$, $AC_{B_{09}}$ is the Afterward Coordinate of band B09, $G_L$ is the granule length. Essential to note is that, given the variations in spatial resolution between bands, the shift values are scaled specifically for each band with the resolution scale factor $\frac{R_{B_k}}{R_{B_{02}}}$, where $R_{B_k}$ and $R_{B_{02}}$ are the spatial resolutions of band $B_k$ and $B02$, respectively. This equation allows us to calculate the Prior Coordinates for each band, facilitating the process of band georeferencing. 

The Afterward coordinates are, instead, prompted using ancillary information by applying the following linear proportion:

\begin{equation}
    G_L : B_{l_{k}} = G_a : \Delta_{B_k}
    \label{eq: geoEq}
\end{equation}

where 
\begin{itemize}
    \item $G_a$ is the granule length expressed in arc of latitude/longitude, prompted from the polygon coordinates at positions 0 and 1 of the granule (Fig. \ref{fig:geo-granule});
    \item $B_{l_{k}}$ is the length of the $k$ band expressed in px;
\end{itemize}
\vspace{0.1in}

Solving the proportion in Equation \ref{eq: geoEq} gives the offset $\Delta_{B_k}$ referred to a generic $k$ band as:

\begin{equation}
\Delta_{B_k} = \frac{B_{l_{k}} \cdot G_a}{G_l}
\end{equation}

In the end, the $AC$ can be calculated as:
\begin{equation}
    AC_{B_k} = PC_{B_k} + \Delta_{B_k}
\end{equation}

The set of AC and PC enables the coarse georeferencing of each pixel within the $k$ spectral band.

After computing the coordinates of the band $B_k = B_{8A}$, it is now possible to crop collate the various \gls{L1C} products and crop them on the area of the band $B_{8A}$. 
To retrieve the same area on the \gls{L1C} products, for each band in $B_S$, we extract and collate the correspondent band in each of the \gls{L1C} tiles that intersect the polygon used to download granules. The resulting mosaic is then cropped on the \gls{AOI} by using the geographical coordinates retrieved by using the \gls{CG} scheme.
Then, the cropped mosaic of the various bands is resampled to match the band with the coarsest resolution. 

Finally, to detect thermal hotspots in \gls{L1C} data, we used a simplified implementation of \cite{massimetti2020volcanic} to generate a hotmap containing thermal hotspots and extract bounding boxes. This algorithm\deleted[id=rev1]{s} relies on the bands $B_S=[B_{8A}, B_{11}, B_{12}]$, which we selected as band collection.
Therefore, we cropped and mosaicked the bands in $B_S$ of the \gls{L1C} products using the coordinates of the georeferenced and coarsely registered raw granules.
To spot thermal hotspots on \gls{L1C} data, as for the original implementation of Massimetti et al. \cite{massimetti2020volcanic}, we derived a hotmap containing candidate anomalies fulfilling four logical conditions that are calculated by applying fixed thresholds on the ratios of \gls{TOA} reflectance of the different bands. 

More specifically, as in the original work \cite{massimetti2020volcanic}\added[id=rev1]{,} the logical equation to activate a pixel $p$ in the hotmap is:
\begin{equation}
    p = \alpha \oplus \beta \oplus S \oplus \gamma
\label{eq: hotmapEq}
\end{equation}
where $\oplus$ is the logical sum between the logical conditions $\alpha$, $\beta$, $S$, and $\gamma$. The latter are calculated as a function \added[id=rev1]{of} the \gls{TOA} reflectance values of the bands $B_{8A}, B_{11}, B_{12}$ ($\rho_{8a}$, $\rho_{11}$, $\rho_{12}$) as follows:

\begin{equation}
\begin{cases}
\alpha =(
\frac{\rho_{12}}{\rho_{11}} \ge 1.4) \otimes (
\frac{\rho_{12}}{\rho_{8A}} 
\ge 1.2)\otimes(\rho_{12} \ge 0.15) \\
\beta ={(
\frac{\rho_{11}}{\rho_{8A}}
\ge 2)\otimes(\rho_{11}\ge 0.5)\otimes(\rho_{12} \ge 0.5)}\\ 
SR \triangleq SUR\{ \alpha \oplus \beta \} \\
S={(\rho_{12} \ge 1.2)\otimes(\rho_{8A} \le 1)\oplus (\rho_{11} \ge 1.5)\otimes(\rho_{8A} \ge 1)} \\
\gamma = (\rho_{12} \ge 1) \otimes (\rho_{11} \ge 1) \otimes (\rho_{8A} \ge 0.5)\otimes SR

\end{cases}
\end{equation}

where $SUR\{\}$ and $\otimes$ are respectively the surrounding and the logical and operation. 

Then, we extracted a bounding box for every cluster of pixels in the hotmap using \textit{scikit-image} \cite{skimage}.
Finally, to minimize the false detection rate, we filtered those bounding boxes surrounding a number of active pixels in the hotmap lower than 9 to discard small clusters of pixels as suggested in \cite{massimetti2020volcanic}. 

In the original implementation, the authors extract\added[id=rev1]{ed} statistical information related to each cluster of pixels to set cluster-dependent thresholds to reduce the false positive rate. We omitted this step in our implementation since we are not interested in pixel-level information. Indeed, the creation of a hotmap is used only for binary classification and bounding box extraction. 

The work by Massimetti et al. \cite{massimetti2020volcanic} was initially implemented for a volcanic eruption. However, other works such as \cite{liu2021detecting} consider volcanic eruptions and wildfires as thermal anomalies using the same Sentinel-2 bands and threshold-based detection methods. 
Given the methodological similarity, we also used the approach of \cite{massimetti2020volcanic} to detect fire events in \gls{L1C} tiles. 

After extracting the bounding boxes on the mosaicked and cropped \gls{L1C} tiles, we reprojected them on the raw products using an affine transformation and manual tuning. 
Indeed, it is essential to note that the bounding boxes extracted from the \gls{L1C} products cannot be straightforwardly used for the raw data due to their different projections. While the raw products are represented in azimuth-range coordinates ($EPSG:4326$), \gls{L1C} products are in latitude-longitude projected in the \textit{WGS84} system. A further issue is added by the geometrical correction applied to the \gls{L1C} images, in which a resampling using a 90m DEM (PlanetDEM 90)\footnote{Reference Document: GMES-GSEG-EOPG-TN-09-0029 (v2.3)} is employed. 
In order to exploit the \gls{L1C} bounding boxes, we warped the points from the \gls{L1C} to the raw coordinates with an affine transformation. 
We defined the transformation matrix from \gls{L1C} to raw coordinates using the coordinates of three corner points. After applying the correction, our results showed that there still exists a discrepancy between the bounding box and the actual event in several cases. In order to rectify the issue, we manually tuned the bounding boxes for eruptions and created a buffer for fire events. 
Finally, we performed a visual inspection to validate \replaced[id=rev1]{each}{ the} presence/absence of events in the \textit{useful granules}. Except for this step, all preceding raw data processing procedures in our methodology can be automated through the \textit{\gls{PyRawS}} package, which offers suitable API functions to implement the described \gls{CSC}, \gls{CG} techniques, \gls{L1C} manipulation, and detected events transfer between \gls{L1C} and Raw data. It also includes the simplified implementation of the Massimetti \cite{massimetti2020volcanic} algorithm previously reported.

An example\deleted[id=rev1]{s} of applications of the different methodology steps by using \textit{\gls{PyRawS}} on raw granules and \gls{L1C} data of THRawS are depicted in Fig. \ref{fig: DatasetsExamples}.

\section{Results}
\label{sec: Results}

\begin{table}[b]
\centering
\caption{Number of events, useful granules, and discarded granules in the dataset.}
\label{tab:ds}
\begin{tabular}{@{}lccc@{}}
\toprule
\textbf{Product/event} & \textbf{Eruptions} & \textbf{Fire} & \textbf{Not-event} \\ 
\midrule
raw products                 & 58 & 20 & 10 \\ 
raw useful granules          & \multicolumn{2}{c}{\hspace{0.15cm}135} & 11 \\ \midrule
Total raw useful granules    & \multicolumn{3}{c}{146} \\ 
Discarded granules           & \multicolumn{3}{c}{707} \\ \midrule
Total number of raw granules & \multicolumn{3}{c}{905} \\ 
\bottomrule
\end{tabular}
\end{table}

\begin{table}[tb]
    \centering
    \caption{Patch study analysis highlighting the number of events and not-events, as well as the proportion of events, as the output of the different patch decomposition process. The values used for the generation of THRawS dataset are marked in bold.}
    \label{tab:patch_study}
            \begin{tabular}{p{0.7cm}p{1.3cm}p{1.4cm}p{1.4cm}p{1.4cm}}
            \toprule
            \textbf{Overlap (\%)} & \textbf{Patch Size} & \textbf{No. of Thermal hotspot samples} & \textbf{No. of Not-Thermal-hotspots samples} & \textbf{Proportion of Events} \\
            \midrule
            \textbf{0.25} & \textbf{(128, 128)} & \textbf{1090} & \textbf{33335} & \textbf{0.031663} \\
            0.25 & (256, 256) & 581 & 6916 & 0.077498 \\
            0.25 & (384, 384) & 527 & 3298 & 0.137778 \\
            0.25 & (512, 512) & 541 & 1907 & 0.220997 \\ \midrule
            0.33 & (128, 128) & 1211 & 37957 & 0.030918 \\
            0.33 & (256, 256) & 783 & 9009 & 0.079963 \\
            0.33 & (384, 384) & 527 & 3298 & 0.137778 \\
            0.33 & (512, 512) & 541 & 1907 & 0.220997 \\ \midrule
            0.50 & (128, 128) & 2158 & 65315 & 0.031983 \\ 
            0.50 & (256, 256) & 1220 & 14080 & 0.079739 \\
            0.50 & (384, 384) & 1102 & 6395 & 0.146992 \\
            0.50 & (512, 512) & 852 & 2973 & 0.222745 \\ \midrule
            0.75 & (128, 128) & 8345 & 248848 & 0.032446 \\
            0.75 & (256, 256) & 4582 & 50651 & 0.082958 \\
            0.75 & (384, 384) & 3326 & 18706 & 0.150962 \\
            0.75 & (512, 512) & 2189 & 7603 & \textbf{0.223550} \\
            \bottomrule
            \end{tabular}
\end{table}

This section provides a quantitative analysis of the THRawS dataset, covering aspects spanning from the composition of its granules to its representation in time and space. 


\subsection{Granule Distribution}
As detailed in Table \ref{tab:ds}, THRawS comprises 88 thermal hotspot events initially selected as 58 volcanic eruptions, 20 fire events, and 10 not-events. 
It is crucial to notice that each of the 88 elements corresponds to one or more \textit{useful granules}, i.e., granules containing thermal hotspots that were appropriately annotated through the proposed approach. Furthermore, eruption events do not necessarily correspond to granules containing volcanic eruptions only but also wildfires.
This is due to using a standard algorithm \added[id=rev1]{by} Massimetti et al. \cite{massimetti2020volcanic} to identify \textit{useful granules} that do not distinguish between volcanic eruptions and wildfires. This issue could be easily solved by manually reviewing the \textit{useful granules}. However, since the goal of THRawS is to create a thermal hotspot dataset and not to differentiate between volcanic or wildfires, this step was not performed. 

Hence, the total number of \textit{useful granules} in the THRawS dataset is 146, of which 135 correspond to thermal anomalies, and 11 granules depict the volcanic areas where no thermal hotspot was observed.
Furthermore, when the discarded granules by the proposed methodology are considered, the total number of granules included in the THRawS dataset is 905. 

\begin{figure}[t]
    \centering
    \includegraphics[scale=0.25]{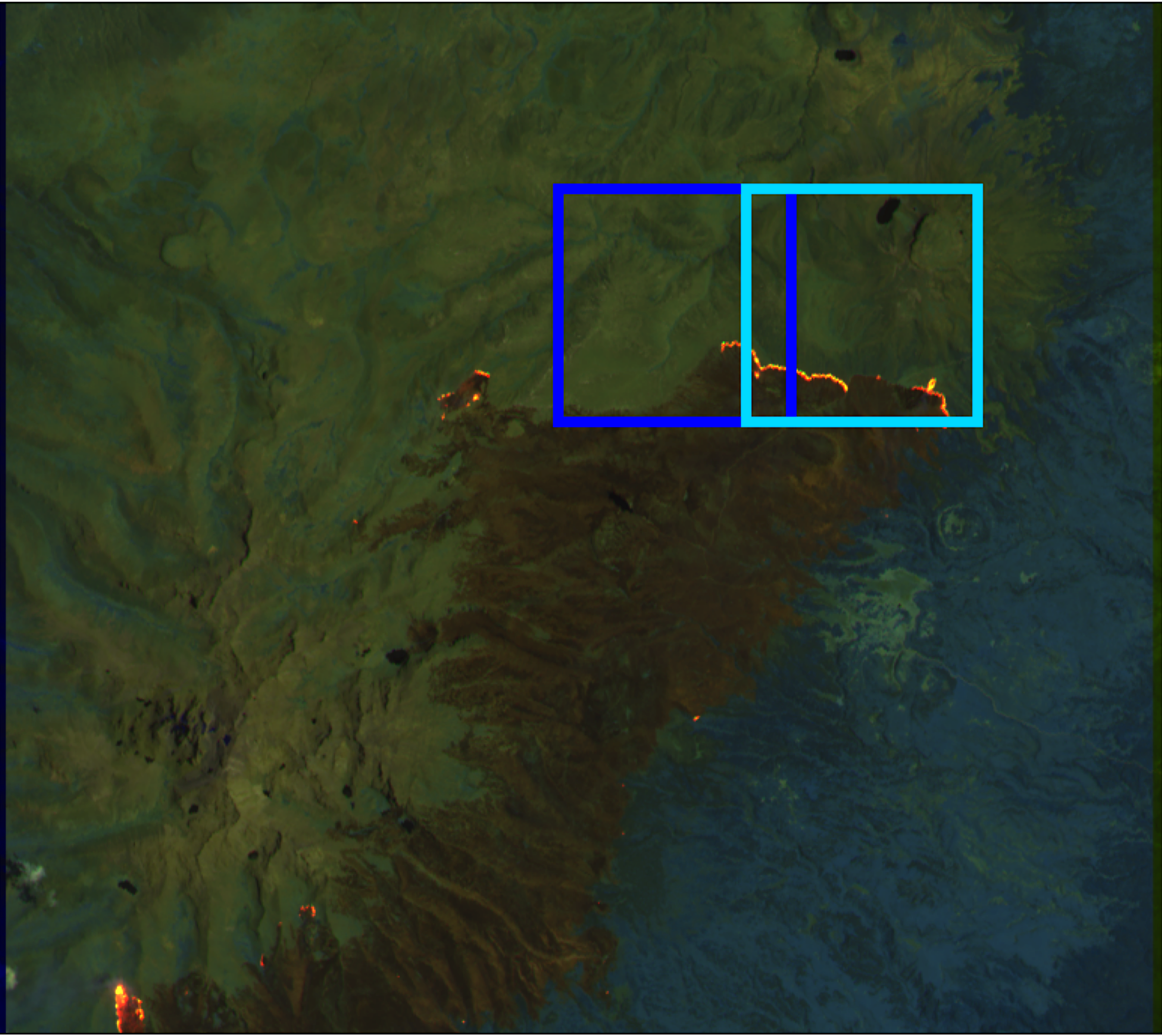}
    \caption{Demosaicing of a raw granule by using 256x256 patches with 25\% of overlap. }
    \label{fig: patchification}
\end{figure}

Notably, the number of images containing \added[id=rev1]{the} event\added[id=rev1]{s} can be significantly increased by cropping the bands $B_S=[B_{8A}, B_{11}, B_{12}]$, whose size is $1152 px \times 1296 px$, into patches of smaller size. 
This step is typically implemented in current smallsats missions to isolate patches containing different spatial features and ensure meeting the onboard memory requirements by limiting the size of patches provided as input to \gls{ML} models \cite{guerrisi2023artificial, GiuffridaPhiSat, derksen2021few, derksen2021few, meoni2024ops}. 
Figure \ref{fig: patchification} illustrates this concept by demonstrating the effect of employing $256 px \times 256 px$ patches with a $25\%$ overlap in raw granule patches decomposition. 

Therefore, the number of dataset samples can be significantly increased depending on the dimensions of the patches and their degree of overlap.
 Table \ref{tab:patch_study} presents a comprehensive analysis of the impact of patch sizes on the THRawS dataset using varying patch sizes and degrees of overlap. 

In the computational methodology employed for this study, a patch is deemed to encompass an event if the spatial extent delineated by its annotation exceeds a threshold of 5 pixels within the patch coverage.
 As shown by Table \ref{tab:patch_study}, decomposing the single granules into patches as requested by the typical onboard hardware requirements leads to a significant increment in the number of samples containing thermal hotspots from hundreds (i.e., number of granules) to several thousand (i.e., number of patches) depending on the overlap and patch size. To generate THRawS, we opted for \deleted[id=rev1]{value uses }25\% overlap and patch sizes of (128x128), leading to more than 1000 hotspot sample patches without significantly increasing the correlation among patches due to large overlap sizes. 

However, applying such a patch decomposition strategy would lead to a high imbalance in training patches between the dataset's thermal hotspot/no-thermal hotspot classes.  Because of that, class imbalance shall be adequately handled by upsampling the event or downsampling the non-event class. 



\subsection{Temporal and Geographical Coverage}
\label{subsec: DatasetCoverage}
\begin{figure}[t]
    \centering
    \includegraphics[scale=0.575]{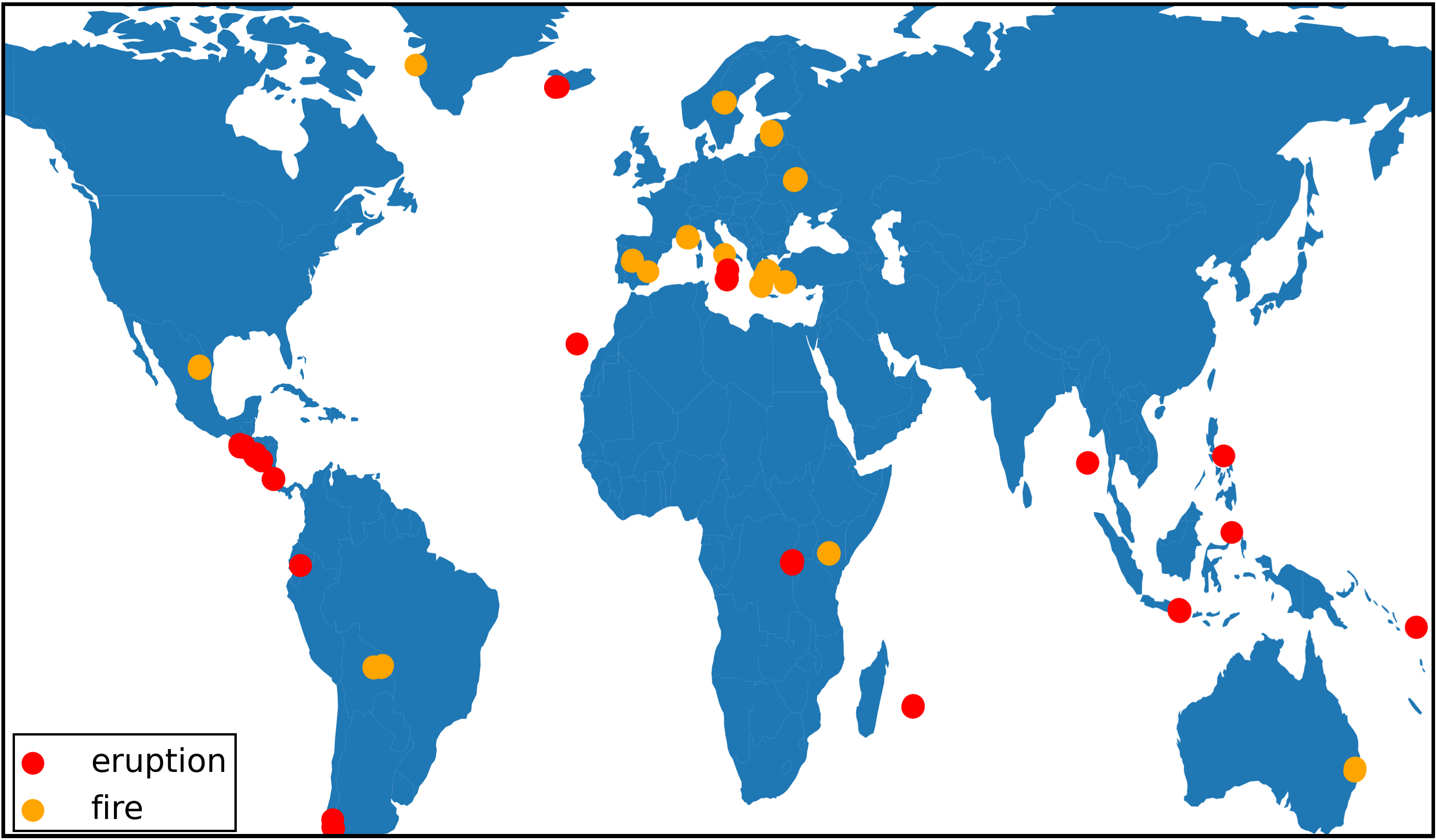}
    \caption{Pictorial view of the geographical distribution of the ``fire" and ``eruption" raw useful granules in the THRawS dataset.}
    \label{fig: GeographicalDistribution}
\end{figure}

Fig. \ref{fig: GeographicalDistribution} shows the geographical distribution of volcanic eruptions and wildfires. 
Volcanic eruptions are primarily concentrated in Central/South America, Africa, Indonesia, the Philippines, and some European islands (such as Sicily, the Canary Islands, and Iceland). 

Wildfires scenes are, instead, concentrated in Europe, Australia, Africa, Greenland, and Central/South America.
When considering both wildfires and volcanic eruptions, the thermal anomalies contained in THRawS have almost global coverage in the various continents except for continental Asia, North America, and Antarctica. 

It is worth noticing that given the nature of thermal hotspots included in the dataset, most of the images focus on rural areas far from urban centers.  
In particular, wildfire hazards are more likely in areas where abundant biomass build-up in the wet season can be converted into fire fuel during the local dry season \cite{abbott2008natural}, such as in tropical countries. Because of that, eventual \gls{AI} models aiming to detect thermal anomalies in urban areas might require additional complementary granules depicting wildfires in such areas. 

For what concerns time representativeness, volcanic eruptions were selected by picking up to three volcanic events for each eruption as defined in the Smithsonian Institution
database \cite{VolcanoDatabaseWeb}. In particular, by picking up the Sentinel-2 granules whose sensing times match the day of the eruption starting date, we included volcanic events in their early phase to enable studies on the early detection of volcanic eruption. Furthermore, given the presence of other events whose sensing times are successive months or years to the first event, THRawS features volcanic events in different stages to enlarge its time representativeness. 
Instead, wildfires in vegetated areas are more likely during the local afternoon since air has warmed and the ground drier than in the morning hours \cite{andela2015new}. 
Given the Sentinel-2 orbit characteristics, this implies that the database will mostly contain more severe and more protracted fire events.
In general, the sensing time of events included in THRawS ranges from 16/01/2017 until 01/06/2022, which spans the early stage of the Sentinel-2 sensor life until the dataset creation time, encompassing possible differences in the Sentinel-2 sensors calibration or mission set-up.

\subsection{Data--Label Representativeness}


We implemented a comprehensive two-step approach to validate the dataset. In the first step, we conducted a preliminary selection process, followed by a thorough visual inspection of the \gls{L1C} data. This step was crucial for ensuring the initial quality and relevance of the data. In the second step, we focused on a detailed visual analysis of the \textit{useful granules}. These granules were specifically chosen by a well-regarded reference algorithm, as detailed in the study by Massimetti et al. \cite{massimetti2020volcanic}. This allowed us to assess the practical utility of the data as defined by established algorithmic criteria. \added[id=rev1]{Visual inspection was performed by two independent experts. No significant errors in labelling warm thermal hotspots arouse from the application of our implementation of the reference algorithm}\cite{massimetti2020volcanic} \added[id=rev1]{on \gls{L1C} products. Any errors in bounding box warping from \gls{L1C} to raw \textit{useful granules} was manually corrected at the best of the authors' possibilities.}

The overarching goal of this two-step validation process was to confirm the accuracy and relevance of the events recorded within the \gls{AOI}. By cross-referencing with existing datasets, we aimed to reduce the likelihood of false positives, which could arise from the inherent limitations or biases of the reference algorithm. This methodical approach enhances the reliability and applicability of our dataset for further analysis and application.

\section{Discussion and future perspectives}
\label{sec: Discussion}
This section discusses the use of raw multispectral imagery for \gls{ML}-based onboard applications. Moreover, it discusses the suitability of the proposed methodology to generate datasets with sufficient fidelity in representing sensory-produced data.

\subsection{Perspectives on usability Raw Data}

As previously mentioned, providing raw imagery is to investigate lightweight onboard pre-processing algorithms or alternative end-to-end solutions to reduce latency and energy consumption, which are of utmost importance for early section systems. 
The idea behind the use of raw data is that \gls{ML} solutions could be trained to handle possible disturbances due to the use of raw imagery, curtailing the need for high-quality data. Some preliminary studies \added[id=rev1]{have }already demonstrated the potential of \gls{DL} models to appropriately 
handle registration errors \cite{fanizza2022transfer, del2023first} by including emulated misregistered data during the training. 

In particular, our parallel work \cite{del2023first} performs a step-ahead by providing a preliminary demonstration of the feasibility of using coarsely registered Sentinel-2 real raw imagery for onboard vessel detection with coarsely registered images.
For this reason, these preliminary results, when combined with the advantageous quality/latency trade-offs of the proposed \gls{CSC} technique used to design THRawS (please refer to Appendix), represent a valuable example on the value of raw data can be used to investigate lightweight energy-efficient trade-offs. A detailed benchmarking of different registration techniques in terms of timing and accuracy on THRawS is provided in\added[id=rev1]{ the }Appendix.

In addition to lightweight processing experimentation, similar\deleted[id=rev1]{ly} to our previous work \cite{del2023first},  the provided THRawS dataset can be a useful tool for the training \gls{ML} models for end-to-end processing onboard thermal hotspot detection and classification on Sentinel-2 raw imagery on rural areas.

However, differently from THRawS, the dataset used in our previous study  \cite{del2023first} was built via manual labelling because of the lack of an automated procedure to handle dataset creation. In view of that, due to the possible upcoming release of further raw Sentinel-2 data, we believe that the presented methodology could play a fundamental role in fostering future research on raw multispectral imagery by significantly curtailing the time for dataset preparation, whose impact on research in terms of development time is already discussed in numerous studies \cite{OPSSATSpecifications, sebastianelli2021automatic}. As previously mentioned, the fundamental steps of the proposed methodology are made reproducible through our dedicated open-source software \textit{\gls{PyRawS}}, which facilitates the handling of raw data significantly.

Hence, the provided methodology could be a useful instrument to start filling the gap \replaced[id=rev1]{in}{of} raw data availability by providing an automated procedure for dataset creation and labelling on Sentinel-2 imagery and beyond.
Indeed, although the present methodology targets Sentinel-2 raw data, its usability is not limited to Sentinel-2 onboard applications, being in principle applicable to other multispectral pushbroom imagery showing similar processing chains \cite{GiuffridaPhiSat,melega2023implementation}, although with opportune modifications.

\subsection{Fidelity in representing Sentinel-2 raw sensory Data}
As mentioned in Sec. \ref{sec: sota}, the proposed raw data are decompressed \gls{L0} data with additional metadata. Because of that, they differ from the data produced by the imager in terms of the presence of such additional metadata and the non-compensated effects due to onboard compression and equalization. 

We used additional metadata only in the phase of dataset preparation (e.g., the coordinates of the granule's corners). However, we do not require metadata information in the granule processing chain, for instance, to perform the proposed \gls{CSC}.
Regarding the near-lossless compression applied on board Sentinel-2 satellites before the data download, the scheme introduces information losses that affect the other products in the processing chain, such as \gls{L1C} data. Since numerous works relying on \gls{ML} for the detection of specific events on \gls{L1C} data exist \cite{di2022early,del2021board, mateo-garcia_towards_2021}, it is possible to conclude this information loss shall not hinder the usability of \gls{ML} on raw data.

From what concerns onboard equalization, it consists of a bilinear function \cite{S2CalibrationPlan} that is applied prior to the wavelet compression to minimize its effects. 
 Assessing the impact of onboard equalization on the performance of a generic \gls{AI} model and comparing Sentinel-2 raw to sensory data without proper experimentation is harsh. In general, however, being natively designed to be performed on board satellites, such algorithms lead to a low computational burden. Because of that, it is reasonable to assume that such algorithms could be part of the onboard preprocessing pipeline in case their presence will be proven to be fundamental to ensure sufficient performance of onboard \gls{ML} models.

\subsection{Limitations of the proposed dataset creation methodology}
\added[id=rev1]{Manual fine-tuning and visual inspection are still required to correctly transfer the bounding boxes from \gls{L1C} to Raw products. This fact currently represents the main obstacle to fully automate the dataset creation procedure and is due to local deformations in the spatial mesh of \gls{L1C} products due to orthorectification and geometric calibration procedures. 
To solve this issue, possible solutions require the matching of local spatial features in the areas of the bounding boxes through key-points identification and matching algorithms, such as SuperGlue} \cite{sarlin2020superglue}\added[id=rev1]{, LightGlue} \cite{lindenberger2023lightglue}\added[id=rev1]{, and others. 
However, a detailed investigation of this problem is beyond the scope of this work and will be investigated in a future study.
}
\subsection{THRawS dataset limitations}

\added[id=rev1]{In our study, we identified several limitations that warrant further investigation. The first pertains to the application of the algorithm by Massimetti et al.} \cite{massimetti2020volcanic} \added[id=rev1]{for detecting thermal hotspots in both volcanic and wildfire events by using uniform threshold value across all events and locations} \cite{liu2021detecting}. \added[id=rev1]{However, our study did not implement the cluster-based thresholds that were integral to the original algorithm by Massimetti et al.}  \cite{massimetti2020volcanic} \added[id=rev1]{to reduce the false positive rates. Our choice is motivated by fact that we do not aim to provide an accurate pixel-level hotmap, but a coarse detection of the events. However, this approach, when combined to our simplified clustering solution and to the buffer created on fire events to mitigate the inaccuracies of L1C/Raw warping, could influence the precision of bounding box placements, necessitating additional tuning for applications that demand exact positioning.}

 \added[id=rev1]{A notable shortcoming in our study stems from the uneven representation of wildfire and volcano events within our dataset. While we have included both wildfires and volcanic eruptions to broaden the spectrum of thermal anomalies, our objective does not encompass the differentiation between these types of thermal anomalies.  Despite that,  we acknowledge that the inclusion of additional wildfire events in North America, Asia, and others with currently limited coverage could significantly benefit our dataset by further enlarging the variety of thermal anomalies included. Moreover, as detailed in Sec. \ref{subsec: DatasetCoverage}, the dataset predominantly features severe and prolonged fire events due to the operational characteristics of Sentinel-2, which may limit the representation of less intense fires commonly found in diverse ecosystems such as savannas and forests. Future studies could consider integrating raw data from additional available sensing platforms to enhance the diversity and representativeness of thermal anomaly events in urban and natural environments.}

 \added[id=rev1]{A dedicated follow-up study will address these limitations by enlarging the population of thermal anomalies and by improving the methodology to select the fire events to enable the delivery of an accurate segmentation mask.} 

\added[id=rev1]{ Finally, it is worth noting that such limitations do not affect the quality of the proposed methodology to automate the design of Raw multispectral data but are, instead, specific to the approach used to select target events on \gls{L1C} for the THRawS dataset, which simply represents a use case of the proposed methodology. 
}

\section{Conclusion}
\label{sec: Conclusions}
This work addressed the lack of available \added[id=rev1]{datasets targeting} raw optical multispectral data for \added[id=rev2]{onboard satellite early detection of target events}  \deleted[id=rev2]{applications} \deleted[id=rev2]{relevant to onboard satellite processing}. To fill this gap, we present the first methodology to automate the creation of dataset containing raw multispectral pushbroom data for target object\deleted[id=rev1]{s}/event\deleted[id=rev1]{s} de\deleted[id=rev2]{s}tection. The methodology was demonstrated on Sentinel-2 raw data for the creation of the THRawS dataset, containing warm thermal hotsposts. THRawS features a collection of elements between volcanic, fire events, and not-events that correspond to more than 1000+ 128x128 patches containing events. To the best of our knowledge, THRawS and the one provided by our previous work \cite{del2021board} are the very first open-source database including raw images from a multispectral pushbroom imager. 
\replaced[id=rev1]{The h}{H}igh degree of reproducibility and automation in the application of our methodology is ensured through our open-source toolbox \textit{\gls{PyRawS}}.

The main motivation behind our work is fostering the research on lightweight pre-processing or end-to-end processing solutions prior to the application of \gls{ML} processing on board satellites.

To this aim, for the creation of THRawS, we adopted a lightweight \gls{CSC} solution for the processing of raw data, which features interesting trade-offs in terms of computational intensity and quality of registration that makes it promising \replaced[id=rev1]{for}{towards} its application on board satellites. 

Future studies will \added[id=rev1]{focus on improving the automation of the proposed methodology and on refining and augmenting the THRawS dataset with a particular emphasis on enhancing the validation approach for label generation. This initiative will involve the rigorous evaluation of label accuracy and the incorporation of advanced techniques to improve the reliability of training data. Critical to this process will be the establishment of robust validation protocols that can effectively handle the inherent complexities and variability of the data sets used. Moreover, we will } investigate the suitability of the raw data to be processed by \gls{ML} on board satellite with minimal pre-processing and compare the obtained trade-offs in terms of application-specific performance and energy efficiency to the current state-of-the-art techniques.



\section{Data availability}
The dataset associated with this study is available and can be accessed through the following DOI: \textit{10.5281/zenodo.7908728}. In addition to the dataset itself, supplementary materials such as code scripts, data dictionaries, or documentation may also be available through the repository to facilitate understanding and replication of the research.

\ifCLASSOPTIONcaptionsoff
  \newpage
\fi

\appendix
\subsection{The Sentinel-2 mission}
The Sentinel-2 mission is part of the Sentinel constellation and 
comprises two identical satellites, \gls{S2-A} and \gls{S2-B}, which provide data respectively since 2015 and 2017. These satellites collectively provide five days of revisit time at the equator, ensuring near-global coverage of land areas, 
the Mediterranean and closed seas, and coastal waters globally.
Each satellite is equipped with a \gls{MSI} sensor that consists of 12 detectors arranged in a staggered array perpendicular to the flight direction of the satellite. 
The detectors capture light reflected from the Earth's surface in narrow strips, or ``swaths" ($290 km$ width \cite{drusch2012sentinel}), as the satellite moves forward, which are then combined to create an entire scene.

Each detector measures the Earth's radiance in 13 \gls{VNIR} and \gls{SWIR} bands with a spatial resolution between $10$ and $60 m$ and acquires data with a pushbroom imaging mode in a \gls{SSO} \cite{richards2022remote, S2PSD}.
A description of Sentinel-2 bands in terms of spatial resolutions, spectral content and resolutions, and possible target applications is provided in Table \ref{tab: S2Bands}. 

\begin{table*}[t]
\centering
\caption{Description of Sentinel-2 bands in terms of spatial and frequency resolution and possible applications \cite{Sentinel2Hub}}
\footnotesize{
\begin{tabular}{p{0.5cm}p{2cm}p{2cm}p{1.2cm}p{9cm}}
\toprule
\textbf{Band} &
  \textbf{Spatial resolution [m]} &
  \textbf{Central Wavelength [nm]} &
  \textbf{Bandwidth [nm]} &
  \textbf{Possible applications} \\ \midrule
B1 &
  60 &
  443 (Ultra Blue) &
  20 &
  Detection of coastal and aerosol \\ \hline
B2 &
  10 &
  490 (Blue) &
  65 &
  Soil and vegetation discrimination, forest mapping, identification of human-made features. \\ \hline
B3 &
  10 &
  560 (Green) &
  35 &
  Detection of oil or vegetation of water surfaces. \\ \hline
B4 &
  10 &
  665 (Red) &
  30 &
  Identification of vegetation types, soils and urban areas. \\ \hline
B5 &
  20 &
  705 (VNIR) &
  15 &
  Identification of vegetation types. \\ \hline
B6 &
  20 &
  740 (VNIR) &
  15 &
  Identification of vegetation types. \\ \hline
B7 &
  20 &
  783 (VNIR) &
  20 &
  Identification of vegetation types. \\ \hline
B8 &
  10 &
  842 (VNIR) &
  115 &
  Mapping shorelines, biomass content and detection of vegetation. \\ \hline
B8A &
  20 &
  865 (VNIR) &
  20 &
  Identification of vegetation types. \\ \hline
B9 &
  60 &
  940 (SWIR) &
  20 &
  Detection of water vapour. \\ \hline
B10 &
  60 &
  1375 (SWIR) &
  30 &
  Cirrus and cloud detection. \\ \hline
B11 &
  20 &
  1610 (SWIR) &
  90 &
  Measuring moisture content of soil and vegetation, differentiating between snow and clouds. \\ \hline
B12 &
  20 &
  2190 (SWIR) &
  180 &
  Measuring moisture content of soil and vegetation, differentiating between snow and clouds. \\ \bottomrule
\end{tabular}}
\vspace{0.1cm}
\label{tab: S2Bands}
\end{table*}

\section{Coarse Spatial Coregistration Technique}
\label{subsec: coarseCoregistration}

As discussed in Section \ref{sec: THRawSCasetStudy}, to design THRawS we leveraged a \gls{CSC} solution that represents an example of effective lightweight processing that can be applied to raw data. As mentioned in Section \ref{sec: Discussion}, the proposed \gls{CSC} was also applied to a parallel study to perform onboard vessel detection on Raw Sentinel-2 data \cite{del2023first}. This section provide details on the methodology used to design the \gls{CSC} algorithm and compare it to other \gls{B2B} alignment solutions in terms of accuracy and latency.

\subsection{CSC design methodology}

Let us define $I$ the vector of band indices as follows:
\begin{equation}
 I \triangleq \{02, 08, 03, 10, 04, 05, 11, 06, 07, 8A, 12, 01, 09\}
\label{eq: Indices}
\end{equation}
$I$ is sorted according to the time delays of the various bands compared to the band $B_{02}$\cite{binet2022accurate}.
Let us consider two bands $B_{I(n)}$ and $B_{I(m)}$, with $B_{I(n)}, B_{I(m)} \in B_S$ where $B_S \triangleq [B_x, B_y, ..., B_z]$ is the band collection of a granule that we want to register spatially.
To perform the \gls{CSC}, we apply a shift along and across the satellite-track $\overline{S_{{B_{I(n)}}\_{B_{I(m)}}}}$ to the band $B_{I(n)}$ measured with respect to the resolution of $B_{I(n)}$. $\overline{S_{{B_{I(n)}}\_{B_{I(m)}}}}$ depends only on the couple of bands $(B_I(n), B_I(m))$, the satellite and the detector number producing that raw data. 

For any coupling of satellite, detector number, and couple of bands, the value of $\overline{S_{{B_{I(n)}}\_{B_{I(m)}}}}$ can be calculated by using the relations in Eq. \ref{eq:coregCoeff}:

\begin{equation}
    \begin{cases}
  \overline{S_{{B_{I(n)}}\_{B_{I(m)}}}} = \sum_{k=m}^{n} N_{I(k+1)\_I(k)} \cdot \frac{R_{B_{I(k+1)}}}{R_{B_{I(n)}}}, & n > m\\
  \overline{S_{{B_{I(n)}}\_{B_{I(m)}}}} = - \overline{S_{{B_{I(m)}}\_{B_{I(n)}}}} \cdot \frac{R_{B_{I(n)}}}{R_{B_{I(m)}}}, & n < m 
    \end{cases}
\label{eq:coregCoeff}
\end{equation}
where  $R_{B_{I(k+1)}}$ is the resolution in m of the band 
$B_{I(k+1)}$, and $N_{I(k+1)\_I(k)}$ is the shift  to apply to the band $B_{I(k)}$ to match the band $B_{I(k+1)}$ measured with respect to the resolution of the band $B_{I(k)}$. 

Given a specific coupling, the shift values $N_{I(k+1)\_I(k)}$ are fixed coefficients that we estimated for each of the band couples $(B_{I(k)}, B_{I(k+1)})$ by performing an analysis on volcanic events and not-events before the filtering of raw data granules. 
To this aim, we ran a \gls{DNN}-based method named SuperGlue \cite{sarlin2020superglue} to extract and match keypoints in a pair of bands $(B_{I(k)}, B_{I(k+1)})$ having the same detector number and used the along-track and across-track distance between the couples of matched key-points to provide an estimation of $N_{I(k+1)\_I(k)}$. More specifically, for a couple of bands $(B_{I(k)}, B_{I(k+1)})$ we proceeded as follows. 

Firstly, to ensure that keypoints that were located in adjacent granules could be matched successfully, we first coupled all the granules used for the study that could be stacked along the satellite track.
Hence, for a given along-track-stacked granule $g$, we picked the couple of bands  $(B_{I(k)-g}, B_{I(k+1)-g})$.
Then, to boost feature presence and distinctiveness without introducing noise in the processing chain, we applied a \gls{CLAHE} \cite{musa2018review} algorithm to the stacked bands $(B_{I(k)-g}, B_{I(k+1)-g})$. 
Then, we applied SuperGlue to extract and match keypoints in the enhanced bands $(B_{I(k)-g}, B_{I(k+1)-g})$ and calculated the average along-track and across-track distance over each couple of matched keypoints. We removed the outliers ($\pm 2\sigma$) and averaged the values for both along and across-track offsets. The obtained result provides a value for $N_{I(k+1)\_I(k)-g}$ for the couple of bands $(B_{I(k)-g}, B_{I(k+1)-g})$ of a couple of stacked granules $g$. Finally, the $N_{I(k+1)\_I(k)}$ is obtained by averaging the $N_{I(k+1)\_I(k)-g}$ over all the couples of stacked granules $g$.
We iterated this process for 13 predefined couples of bands.  

Since no granule among the volcano events was generated by detector number 6, to measure $N_{I(k+1)\_I(k)}$, we used the database inclusive of not-events. 

However, we found that band $B_{10}$ was difficult to correlate with the bands $B_{03}$ and $B_{04}$ because of the significant difference in the spectral content. This makes the estimated values of $N_{04\_10}$ and $N_{10\_03}$ unreliable. 
To solve this problem, we extracted two additional coefficients $N_{10\_09}$ and $N_{10\_03}$ and used them to invert Eq. \ref{eq:coregCoeff} and provide better estimates for $N_{04\_10}$ and $N_{10\_03}$.
The extraction of $N_{10\_09}$ was possible because of the higher correlation of the spectral contents of the bands 
 $B_{10}$ and $B_{09}$. 
 
 Regarding settings, we used the Superpoint Neural Network as a feature extractor with outdoor weights as in the original implementation \cite{sarlin2020superglue}.
 To increase the detection and matching capabilities, we tuned the hyper-parameters for our problem, and the best ones have been reported in our source code\footnote{Pyraws matching config: \url{https://shorturl.at/klJL7}; last accessed on 2023-12-19.}. In particular, we increased the matching threshold values to increase the number of strongly matched keypoints, i.e., those having higher probabilities of being the same in both bands.

Concerning the accuracy of the proposed method, Table \ref{tab:CoregError} is a good indicator presenting the mean and standard deviation of the across and along-track offsets for the $B_{8A}-B_{11}$ spectral bands. Such values were obtained by measuring the error due to the proposed \gls{CSC} method with respect to the shifts that one can obtain using the SuperGlue method.  Notably, the standard deviation values exhibit a general trend of being below one pixel for both along- and across-track directions. Although this behaviour has been reported for only the $B_{8A}-B_{11}$ spectral bands, it is essential to note that this trend has been consistently verified for each spectral band pair of the Sentinel-2 images. Also, note that the results presented in the table have been differentiated based on the satellite and detector number. 
This differentiation is necessary due to the empirical observation of dependence on the satellite. 
Furthermore, an additional interesting pattern that is observable in the results is a change in sign when transitioning from odd to even detectors. This pattern is attributed to the fact that the detectors are inverted on the satellite, resulting in an opposite offset sign for even detectors compared to odd detectors. 

It is worth noticing that the reported errors are obtained by using the shift values $N_{I(k+1)\_I(k)}$ matching the average values on the volcanic events before being filtered, which include granules having an acquisition date ranging from 2017 to 2022.

However, in a hypothetical mission scenario, it is possible to recalculate $N_{I(k+1)\_I(k)}$ periodically to optimize them towards specific calibration settings or other possible temporary sensor set-ups. 

\begin{table}[tb]
\centering
\caption{Mean and standard deviation (along- and across-track) of offsets values for each detector of Sentinel-2 optical imagers when performig the registration of $B_{8A}-B_{11}$ spectral bands. Here in the offset is provided in pixel.}
\label{tab:CoregError}
\resizebox{\columnwidth}{!}{%
\begin{tabular}{lllll}
\toprule
 & \multicolumn{2}{c}{\textbf{Sentinel-2A}} & \multicolumn{2}{c}{\textbf{Sentinel-2B}} \\
\textbf{D} & \multicolumn{1}{c}{\textbf{$\mu$}} & \multicolumn{1}{c}{\textbf{$\sigma$}} & \multicolumn{1}{c}{\textbf{$\mu$}} & \multicolumn{1}{c}{\textbf{$\sigma$}} \\ \midrule
1 & {[}-174.8, -1.93{]} & {[}1.01, 0.26{]} & {[}-178.33, -12.78{]} & {[}1.12, 0.44{]} \\ 
2 & {[}188.0, -6.0{]} & {[}0.76, 0.0{]} & {[}186.83, -16.5{]} & {[}1.47, 0.55{]} \\  
3 & {[}-173.08, -2.0{]} & {[}0.79, 0.0{]} & {[}-174.77, -13.0{]} & {[}1.42, 0.0{]} \\  
4 & {[}186.0, -3.2{]} & {[}0.94, 0.42{]} & {[}183.22, -14.06{]} & {[}0.94, 0.24{]} \\  
5 & {[}-170.5, -1.88{]} & {[}0.53, 0.35{]} & {[}-172.5, -13.0{]} & {[}0.62, 0.0{]} \\  
6 & {[}184.0, -2.0{]} & {[}0.0, 0.0{]} & {[}183.0, -12.75{]} & {[}0.82, 0.5{]} \\  
7 & {[}-170.0, -1.0{]} & {[}0.0, 0.0{]} & {[}-173.0, -11.67{]} & {[}0.0, 0.58{]} \\  
8 & {[}185.56, -0.64{]} & {[}1.12, 0.49{]} & {[}184.12, -10.96{]} & {[}0.83, 0.2{]} \\  
9 & {[}-170.55, -1.73{]} & {[}1.04, 0.47{]} & {[}-172.6, -11.0{]} & {[}0.82, 0.0{]} \\  
10 & {[}187.0, 1.5{]} & {[}0.0, 0.58{]} & {[}185.13, -9.33{]} & {[}0.83, 0.49{]} \\  
11 & {[}-176.0, -0.5{]} & {[}0.0, 0.71{]} & {[}-174.88, -11.88{]} & {[}1.13, 0.35{]} \\  
12 & {[}193.0, 4.0{]} & {[}0.0, 0.0{]} & {[}189.17, -7.33{]} & {[}0.41, 0.52{]} \\ \bottomrule
\end{tabular}%
}
\end{table}

\subsection{Quality of \gls{CSC}}
\label{subsec: CoregQualityResults}
In this section, we assess the usability of the proposed \gls{CSC} for onboard satellite processing by comparing it to a reference algorithm called \textit{SuperGlue} \cite{sarlin2020superglue} in terms of quality of registration. 
 Such a study was performed across the entire post-processed THRawS dataset.
This comparison is performed by applying SuperGlue and the proposed \gls{CSC} solution for the alignment of the bands $B_{8A}$ and $B_{11}$. 
As shown in Table \ref{tab:CoregError}, 17 of 24 cases feature a subpixel error both along and across-track compared to the. In 3 cases, the coregistration error is null. The maximum along-track and across-track errors are 1.47 px and 0.71 px, respectively. Such values are significantly lower than the smallest thermal anomaly in the THRawS dataset, which was selected using the nine contiguous pixel criterion implemented by Massimetti et al. to suppress false positives \cite{massimetti2020volcanic}. Therefore, no warm temperature hotspot is misaligned due to the co-registration error.

These results showcase the potential of the proposed \gls{CSC} algorithm, not for its precision in \gls{B2B} alignment but for its energy efficiency and simplification of the problem. The primary aim of \gls{CSC} is to correct misalignments rather than performing complex warping. We juxtapose the \gls{CSC} method with other techniques, noting that while others rely on feature detection and matching, \gls{CSC} efficiently bypasses these steps. 

\subsection{\gls{CSC} Timing Performance}
\label{subsec: CoregResults}

This subsection compares the proposed \gls{CSC} technique in terms of timing performance to SuperGlue, and other coregistration solutions including  the traditional SIFT implementation \cite{lowe2004distinctive}, LightGlue \cite{lindenberger2023lightglue} method, covering different feature descriptors, including Superpoint, ALIKED \cite{zhao2023aliked}, and DISK \cite{tyszkiewicz2020disk}. Differently from the described \gls{CSC}, these techniques are feature-based, i.e., they extract and match keypoints from the bands to be aligned enabling a fine \gls{B2B} alignment.
Results shown in Table \ref{tab:profiling} showcase the profiling of these methods.
This comparison was conducted across the granules of the THRawS dataset, highlighting the practical advantages of \gls{CSC} in onboard satellite processing contexts.
More specifically, we set up an experiment to compare the time to register 10 raw data granules. 
\begin{table}[b]
    \caption{Evaluation of timing performance for SuperGlue, LightGlue (ALIKED, DISK, SIFT, Superpoint), and traditional SIFT-based method in spatial coregistration compared to the developed CSC approach, executed on CPU (Intel® Xeon® Gold 6248 CPU @ 2.50GHz) and GPU (NVIDIA A40-48C with Driver Version 525.105.17 and CUDA Version 12.0).}
    \centering
        \begin{tabular}{lrr}
        \toprule
        \textbf{Method} & \textbf{Average CPU [ms]} & \textbf{Average GPU [ms]} \\
        \midrule
        SuperGlue (Superpoint) & 5167.45 & 1706.30 \\
        Lightglue (ALIKED) & 5651.29 & N/A \\
        Lightglue (DISK) & 7169.25 & 536.51 \\
        Lightglue (SIFT) & 1916.15 & 1225.58 \\
        Lightglue (Superpoint) & 6660.91 & 458.92 \\
        SIFT & 23849.50 & 2189.06 \\
        CSC & \textbf{16.63} & \textbf{1.65} \\
        \bottomrule
        \end{tabular}
    \label{tab:profiling}
\end{table}
For all the aforementioned methods, we measured the time to register the bands $B_{8A}, B_{11}, B_{12}$.  
We profiled both the methods by using PyTorch Profiler \cite{PytorchProfiler} both on an Intel® Xeon® Gold 6248 CPU @ 2.50GHz \gls{CPU} and an NVIDIA A40-48C \gls{GPU} to test the dependency on the time performance on the hardware device. 
Each test case was run three times, and time performance was measured by averaging the results on the three tests. All the tests were performed continuously with 15 warm-up cycles at the beginning of the test series with one raw data granule. 

Concerning the \gls{CSC}, precalculated shifts were stored in memory before running the tests to reduce the overhead time due to the storage memory access. When the other methods were used, shift values were retrieved using the approach and the model set-up described in Section \ref{subsec: coarseCoregistration}. 

On both devices, all the methods show a linear increment of the time to perform the band registration with the number of images. In general, on both \gls{CPU} and \gls{GPU}, our method outperforms the other feature-matching techniques.
In particular, for a single granule with bands $B_{8A}, B_{11}, B_{12}$ our methodology requires 17ms and
1.65 ms on \gls{CPU} and \gls{GPU}, respectively.

\bibliographystyle{IEEEtran}
\bibliography{end2end.bib}







\end{document}